\documentclass[journal]{IEEEtran}
\usepackage{graphicx}
\usepackage{float}
\usepackage{subcaption}
\usepackage{amsmath}
\usepackage{bm}
\usepackage{upgreek}
\usepackage{cite}
\usepackage[numbers,sort&compress]{natbib}
\usepackage{color}
\ifCLASSINFOpdf
\else
\fi
\begin{document}

\title{A GOA-based Fault-tolerant Trajectory Tracking Control for an Underwater Vehicle of Multi-thruster System without Actuator Saturation}
%
%
%

\author{Danjie~Zhu,~\IEEEmembership{Member,~IEEE},~Lei~Wang,~Hua~Zhang~and~Simon~X.~Yang,~\IEEEmembership{Senior Member,~IEEE}
\thanks{This work is supported by the Natural Sciences and Engineering Research Council (NSERC) and the National Key Project of Research and Development Program of China.
Danjie Zhu and Simon X. Yang are with Advanced Robotics and Intelligent System (ARIS) Laboratory, School of Engineering, University of Guelph, Guelph, ON. N1G2W1, Canada (e-mail:\{danjie; syang\}@uoguelph.ca); Lei Wang and Hua Zhang are with the Underwater Engineering Institute, China Ship Scientific Research Center, Wuxi 214082, China (e-mail: {13771115826@163.com; zhanghua702}@126.com).}
}

%
%

\markboth{A GOA-based Trajectory Tracking FTC}%
{Shell \MakeLowercase{\textit{et al.}}: Bare Demo of IEEEtran.cls for IEEE Journals}
%



\maketitle

\begin{abstract}
This paper proposes an intelligent fault-tolerant control (FTC) strategy to tackle the trajectory tracking problem of an underwater vehicle (UV) under thruster damage (power loss) cases and meanwhile resolve the actuator saturation brought by the vehicle's physical constraints. In the proposed control strategy, the trajectory tracking component is formed by a refined backstepping algorithm that controls the velocity variation and a sliding mode control deducts the torque/force outputs; the fault-tolerant component is established based on a Grasshopper Optimization Algorithm (GOA), which provides fast convergence speed as well as satisfactory accuracy of deducting optimized reallocation of the thruster forces to compensate for the power loss in different fault cases. Simulations with or without environmental perturbations under different fault cases and comparisons to other traditional FTCs are presented, thus verifying the effectiveness and robustness of the proposed GOA-based fault-tolerant trajectory tracking design. 

Note to Practitioners:
This paper is motivated by the actuator saturation problem that exists in the trajectory tracking of an underwater vehicle (UV) when encountering power loss of the thruster system. The fault-tolerance trajectory tracking performance is affected by physical constraints of the vehicle when using the traditional methods as they may deduct excessive kinematic/dynamic requirements during the control process, thus inducing the deviation of the tracking trajectory. Therefore, the refined backstepping as well as the grasshopper optimization (GOA) are combined to eliminate the excess, where the refined backstepping is used to alleviate the speed jumps (kinematic outputs) and the GOA is to control the propulsion forces (dynamic outputs) when facing thruster fault cases. This innovates the industrial practitioners that the control design of the vehicle can be improved to avoid the tracking deviation brought by unsatisfied driving commands under fault cases through embedding optimization algorithms. Moreover, for the specific type of UV studied in this paper used for dam detection, simulations regarding practical dam detection such as the 3D polygonal line trajectory tracking and the frequently occurring UV single-fault cases are chosen, which can serve as references for practitioners working in the related field. In the future, underwater experiments of the UV will be investigated, with more effects of the practical environment involved. 
\end{abstract}


%
\IEEEpeerreviewmaketitle

\section{Introduction}
\IEEEPARstart
{T}{he} study of vehicle controls has been extended to various conditions where fault tolerance is widely involved in the corresponding control design, which is denoted as fault-tolerant control (FTC) \cite{Baran2018,Chaosteerwire2019}. Scientists have worked on the FTC study for decades in various fields such as crafts in the air or space, land vehicles and industrial manufacturing \cite{xin2014,Lu2016,TITS_mao2020,Meyer2018,xiong2018,gong2018tec1,motorsCST1, Lang2021}. In previous studies, the FTC is usually applied to alleviate abrupt errors and provides the most feasible solution when inevitable damages happen for equipment in different fields \cite{Cao2021}. However, the research regarding the FTC on underwater vehicles (UV) has not been thoroughly investigated, due to the complexity brought by the underwater environment and the UV system \cite{Seto2020,ocftc1,TITS1_MSV2021,UUV1}.

Corresponding studies on the FTC have been proposed in this century \cite{xin2014,TSMC1,Lu2016,Meyer2018}. Based on these studies, the design of the excessive number of thrusters compared to the number of degrees of freedom (DOF) is raised and accepted as a resolution to the UV FTC problem, which is called thruster reconfiguration \cite{martynova2020,ocftc2}. For example, when unexpected fault cases of the vehicle thrusters occur, the thrusters installed on the vehicle that exceeds the number of DOFs (six: surge, sway, heave, row, pitch and yaw) have enough flexible space to be regulated to provide the required propulsion at corresponding DOFs. To implement the thruster configuration theory in practical cases, the weighted pseudo-inverse matrix method has been proposed, where the fault cases are quantified as degrees of damage and serve as the inputs to form the thruster configuration model \cite{Omerdic2004}. By this method, the process of the FTC is largely simplified, as the required thruster propulsion can be deducted directly through a weighted pseudo-inverse matrix model. Nevertheless, physical constraints of the thruster outputs are rarely considered, thus inducing the over-actuated vehicle issue \cite{Podder2000,Zhang2021}. Additionally, among these studies, most of them work on eliminating the static errors induced by the fault cases. While in UV practical application, the realization of dynamic control on the vehicle's outputs in a real-time manner, which commonly refers to the trajectory tracking control for underwater vehicles, is of crucial importance \cite{Shen2018,TSMC2,UUV2}. 

Therefore, motivated by the over-actuated issue and meanwhile realizing robustness for underwater vehicle trajectory tracking, optimization methods are combined with the tracking control to optimize the vehicle's dynamic outputs within allowable domains during the tracking procedure when encountering fault cases. Among the current mainstream optimization algorithms, the genetic algorithm consumes a long time on iteration, which is not ideal for the UV FTC that requires both fast and feasible solutions \cite{Sinha2020,mohanty2020,dunweitec2}. The neural network is demanding on the choice of data inputs while the UV cannot provide when encountering fault cases, which shares the same concern with the greedy algorithm, as the greedy algorithm needs to decompose the data for processing \cite{dataconvNNChi2019,NNAsmara2020,greedyHul2017}. Hence the swarm intelligence algorithm for optimization stands out to be a preferable method to tackle the FTC application of UVs due to its flexibility of data inputs and fast convergence speed \cite{Guo2018,Celtek2020,TITSwu2021_pso}. Zhu's group has applied Particle Swarm Optimization (PSO) based FTC on the unmanned underwater vehicle, though satisfactory torque outputs are achieved, the traditional PSO method shows poor real-time feedback, which does not conform to the online requirement of UV FTCs \cite{Liu2011,Guo2015}. In this study, an advanced swarm intelligent method named Grasshopper Optimization Algorithm (GOA) is chosen based on its satisfactory balance between fast convergence and accuracy of obtaining optimization results as well as its simple implementation \cite{Saremi2017}. The fast convergence is realized by its simply updated iteration derived from the position of each search agent, which dramatically promotes optimizing efficiency and offers the possibility of real-time feedback for the UV FTC \cite{Guo2021,Xia2021}. At the same time, GOA can provide accurate fault-tolerant results within acceptable driving constraints through the limitations embedded in the optimization algorithm, thus performing adaptive in resolving the over-actuated issue of the UV FTC \cite{jianfa2017,Mishra2020}. 

The contribution of the control strategy proposed in this paper is to combine an advanced swarm intelligence algorithm (GOA) with the fault-tolerant trajectory tracking control to resolve fault cases of a progressed UV without actuator saturation. By identifying and quantifying the degree of damage (power loss) of the multi-thruster system and then efficiently reallocating their forces through the GOA method, a feasible solution with satisfactory accuracy will be given in a fast convergence manner. The strategy establishes a systematic fault identification as well as efficient error elimination process for the UV with abrupt damages; and it first realizes the application of the GOA method in FTC of specific underwater vehicles with a multi-thruster system. Conventional methods such as the constrained control allocation method developed by Durham are based on the basic linear algebra concepts and a means to determine the bounding surface of the attainable moment space, yet the bounding surface is difficult to be addressed \cite{Durham1993}. Commonly applied methods used for constructing UV FTC such as T-approximation or S-approximation cannot thoroughly resolve the problem of actuator saturation due to the inevitable errors brought by the vehicle constraints \cite{Liu2011}. Therefore, the fast convergence speed and ideal accuracy of the GOA method help to amend the commands given by the dynamic controller in time, which accomplishes real-time FTC on the UV trajectory tracking \cite{Saremi2017}.

The rest of the paper is organized as follows. First, the kinematic and dynamic models of an advanced UV, "YuLong", are introduced and the torque-force transition/normalization is defined based on the UV thruster configuration. Next, the fault-tolerant trajectory tracking problem of the UV is described, with its restrictions explained. The grasshopper optimization-based fault-tolerant trajectory tracking control is then proposed, where a refined backstepping algorithm is applied to form the kinematic control component; a sliding mode control works as the dynamic control component; and the grasshopper optimization algorithm is used to form the fault-tolerant component by reconfiguring the thruster force outputs to eliminate the errors produced by different damage degrees of the thrusters in fault cases. In the last section, the effectiveness of the proposed grasshopper optimization-based FTC is evaluated by tracking desired polygonal line or helix trajectories, under the condition that one thruster (single-fault) or two thrusters (double-fault) are supposed to be damaged.

 
\section{UV MODELS AND PROBLEM STATEMENT}
In this section, a typical type of UV named "YuLong" is studied. Its robot models and trajectory tracking problem descriptions are given in the form of specific equations. 
\subsection{Models of the "YuLong" UV}
In this section, a typical type of UV named "YuLong" is studied. Its robot models and fault-tolerant trajectory tracking problem descriptions are given in the form of specific equations. 
\subsubsection{Kinematic and Dynamic Model}
"YuLong" UV is one of the latest UV for dam detection, designed by the Underwater Engineering Institute of China Ship Scientific Research Center, whose diving depth reaches 3000m. Its rough sketch is shown in Fig. 1 and its multi-thruster system structure is shown in Fig. 2. As a UV specially designed for detecting and maintaining a satisfactory condition of the dam in the deep-water area, the robustness and efficiency of the vehicle's operation are crucial for dam detectors.

Among the six degrees of freedom (DOF) of the UV, surge, sway, heave, roll, pitch and yaw, roll and pitch can be neglected because these two DOFs barely have an influence on the underwater vehicle during practical navigation (Fig. 1). Therefore, when establishing the trajectory tracking model to keep a controllable operation of the UV, usually four DOFs surge, sway, heave and yaw are involved. Based on the four involved DOFs, for the kinematic equation of the UV, the velocity vector \textbf{v} can be transformed into the time derivative of the 
trajectory vector $\mathbf{\dot{p}}$ as
\begin{equation}
       \mathbf{\dot{p}}= 
 \begin{bmatrix}
   \dot{x} \\
   \dot{y} \\
   \dot{z} \\
   \dot{\psi}
  \end{bmatrix}
=\mathbf{J}(\mathbf{p})\mathbf{v}=
=\begin{bmatrix}
   \cos{\psi}& -\sin{\psi} & 0 & 0 \\
   \sin{\psi}& \cos{\psi} & 0 & 0 \\
   0 & 0 & 1 & 0 \\
   0 & 0 & 0 & 1
  \end{bmatrix}
  \begin{bmatrix}
   u \\
   v \\
   w \\
   r
  \end{bmatrix},
\end{equation}
where $\mathbf{J}$ is a transformation matrix derived from the physical structure of the UV body, $[u~v~w~r]^T$ represents the velocities at the chosen four axes of the UV (see Fig. 1).
\begin{figure}[h]
\begin{center}
        \includegraphics[scale=0.6]{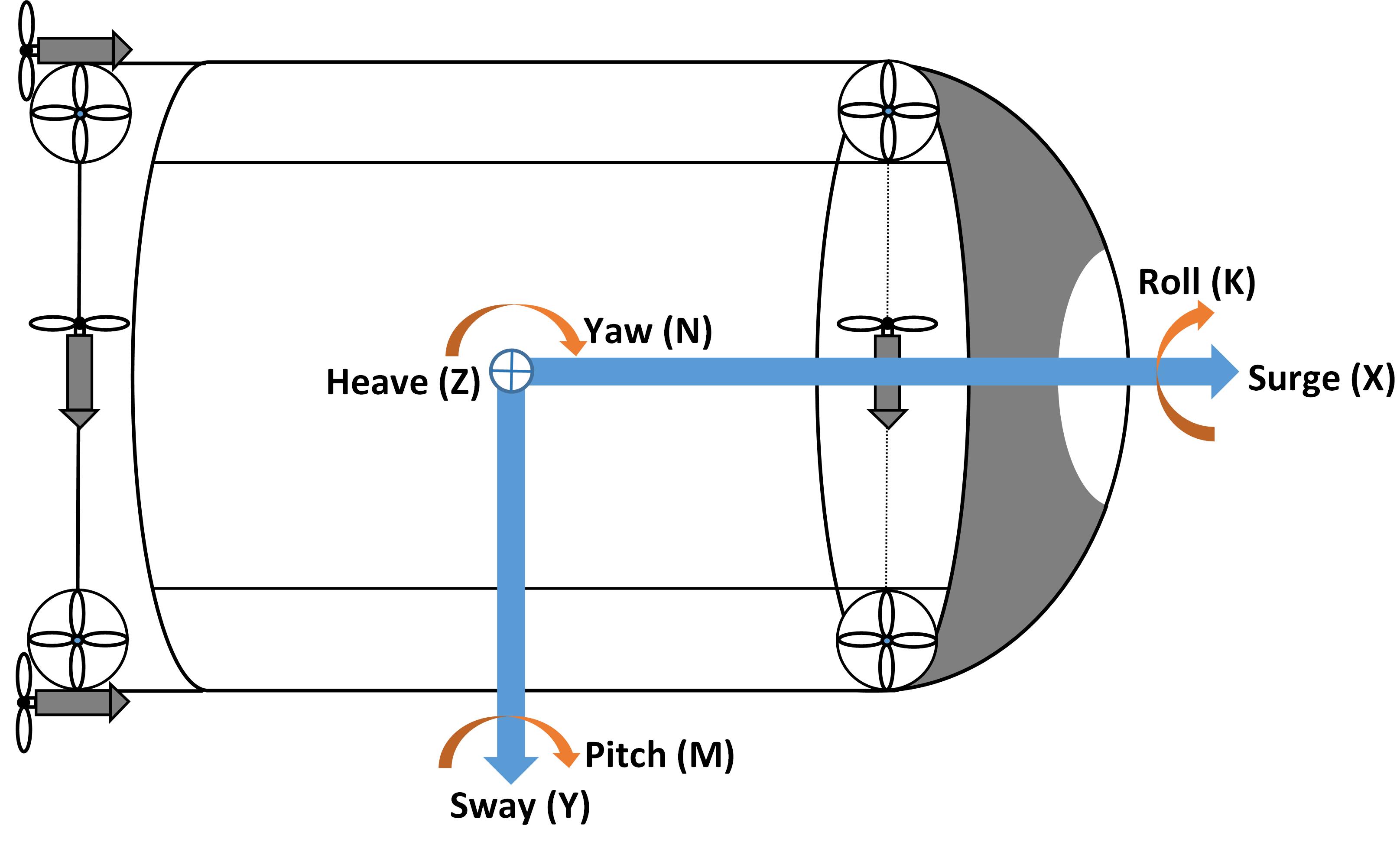}                         
        \caption*{Fig. 1. The reference frame and six degrees of freedom (x, y, z, k, m and n) of the "YuLong" UV (Top view).}	
\end{center} 
\end{figure}

In an actual UV system, several complex and nonlinear forces such as hydrodynamic drag, damping, lift forces, Coriolis and centripetal forces, gravity and buoyancy forces, thruster forces, and environmental disturbances are acting on the vehicle. Considering the origins and effect of the forces, a general dynamic model can be written as
\begin{equation}
\mathbf{M}\mathbf{\dot{v}}+\mathbf{C}(\mathbf{v})\mathbf{v}+\mathbf{D}(\mathbf{v})\mathbf{v}+\mathbf{g}(\mathbf{p})=\boldsymbol{\uptau}\,,
\end{equation}
where M is the inertia matrix of the summation of rigid body and added mass; $\mathbf{C}(\mathbf{v})$ is the Coriolis and centripetal matrix of the summation of rigid body and added mass; $\mathbf{D}(\mathbf{v})$ is the quadratic and linear drag matrix; $\mathbf{g}(\mathbf{p})$ is the matrix of gravity and buoyancy; and $\boldsymbol{\uptau}$ is the torque vector of the thruster inputs. 
As mentioned in the previous section, in this study only four states are considered for the specific model YuLong UV. The torque vector of the thruster input is represented by
\begin{equation}
    \boldsymbol{\uptau}=
    \begin{bmatrix}
       \tau_x & \tau_y &
       \tau_z &
       \tau_n
    \end{bmatrix},
\end{equation}
where $x$, $y$ and $z$ represent the linear displacements of the UV at surge, sway and heave directions, while $n$ represents the angular displacement of the UV at yaw direction (see Fig. 1). 

For the "Yulong" UV, the following parameter values are assigned: inertia matrix $\mathbf{M}=[42~~0~~0~~0;~~0~~153~~0~~0;~~0~~0~~141~~0;~~0~~0~~0~~100]^T$; Coriolis and centripetal matrix $\mathbf{C}(\mathbf{v})=\mathbf{J}^{-1} \mathbf{M} \dot{\mathbf{J}} \mathbf{J}^{-1}$, where $\mathbf{J}^{-1}$ represents the inverse matrix of $\mathbf{J}$ and $\dot{\mathbf{J}}$ represents the derivative of $\mathbf{J}$ and quadratic and linear drag matrix $\mathbf{D}(\mathbf{v})=[42+69u~~0~~0~~0;~~0 ~~319+245v~~0~~0;~~0~~0~~272+86w~~0;~~0~~0~~0~~33+4r]^T$. In addition, the gravity force applied on the vehicle is balanced off by the buoyancy force when the whole system sustains at an equilibrium status.

\subsubsection{Torque-force Transition and Normalization}
According to the physical structure of the "Yulong" vehicle propulsion system (see Fig. 2), the relation between its torque vector and the forces of the thrusters is
\begin{equation}
\boldsymbol{\uptau}=
    \begin{bmatrix}
       \tau_x\\
       \tau_y\\
       \tau_z\\
       \tau_n
    \end{bmatrix}=
    \begin{bmatrix}
       T_1+T_2\\
       T_7+T_8\\
       T_3+T_4+T_5+T_6\\
       T_1+1.4\times T_8-T_2-1.4\times T_7
    \end{bmatrix},
\end{equation}
where $T_1$, $T_2$, $T_3$, $T_4$, $T_5$, $T_6$, $T_7$, $T_8$ are the forces produced by the eight thrusters installed around the vehicle body. 

The eight thrusters are of the same type and are supposed to have the same maximum force $T_m$. Therefore the maximum of the torque vector $\boldsymbol{\uptau_m}$ can be deducted based on Eq. (4) as
\begin{equation}
    \boldsymbol{\uptau_m}=
    \begin{bmatrix}
       \tau_{xm}\\
       \tau_{ym}\\
       \tau_{zm}\\
       \tau_{nm}
    \end{bmatrix}=
    \begin{bmatrix}
       2T_m\\
       2T_m\\
       4T_m\\
       (2+2.8)T_m\\
    \end{bmatrix}.
\end{equation}

On the basis of Eq. (4), divide both sides of Eq. (5) by the maximum torques to restrict the output in a certain range of -1 to 1, and set $\overline{\boldsymbol{\uptau}}=\boldsymbol{\uptau}/\boldsymbol{\uptau_m}$, $\mathbf{\overline{T}}=\mathbf{T}/\mathbf{T_m}$, the vector is transformed into
\begin{equation}
\begin{bmatrix}
       \overline{\uptau}_x\\
       \overline{\uptau}_y\\
       \overline{\uptau}_z\\
       \overline{\uptau}_n
    \end{bmatrix}
    =
\begin{bmatrix}
       \tau_x/\tau_{xm}\\
       \tau_y/\tau_{ym}\\
       \tau_z/\tau_{zm}\\
       \tau_n/\tau_{nm}
    \end{bmatrix}\\ \notag
\end{equation}

\begin{equation}
    =
    \begin{bmatrix}
       \frac{1}{2} & \frac{1}{2} & 0 & 0 & 0 & 0 & 0 & 0\\
       0 & 0 & 0 & 0 & 0 & 0 & \frac{1}{2} & \frac{1}{2}\\
       0 & 0 & \frac{1}{4} & \frac{1}{4} & \frac{1}{4} & \frac{1}{4} & 0 & 0\\
       \frac{5}{24} & -\frac{5}{24} & 0 & 0 & 0 & 0 & -\frac{7}{24} & \frac{7}{24}
    \end{bmatrix}
    \begin{bmatrix}
       T_1/T_m\\
       T_2/T_m\\
       T_3/T_m\\
       T_4/T_m\\
       T_5/T_m\\
       T_6/T_m\\
       T_7/T_m\\
       T_8/T_m
    \end{bmatrix}\\ \notag
\end{equation}
\begin{equation}
    =
    \overline{\mathbf{B}} \begin{bmatrix}
       T_1/T_m\\
       T_2/T_m\\
       T_3/T_m\\
       T_4/T_m\\
       T_5/T_m\\
       T_6/T_m\\
       T_7/T_m\\
       T_8/T_m
    \end{bmatrix}
    =
    \overline{\mathbf{B}} \begin{bmatrix}
       \overline{T}_1\\
       \overline{T}_2\\
       \overline{T}_3\\
       \overline{T}_4\\
       \overline{T}_5\\
       \overline{T}_6\\
       \overline{T}_7\\
       \overline{T}_8
    \end{bmatrix}.
\end{equation}

The above equation has the compact form as
\begin{gather}
\overline{\boldsymbol{\uptau}}=\overline{\mathbf{B}}\,\mathbf{\overline{T}}\notag
\\
\,\mathbf{\overline{T}}=\overline{\mathbf{B}}^{-1}\, \overline{\boldsymbol{\uptau}}\,,
\end{gather}
where $\overline{\mathbf{B}}^{-1}$ is the generalized inverse matrix of $\overline{\mathbf{B}}$.

Therefore, the transition between thruster forces and torques is achieved and normalized. For all the torques and forces in Eq. (6), they are ranged from -1 to 1 ($-1 \leq \overline{\boldsymbol{\uptau}} \leq 1$ and $-1 \leq \overline{\mathbf{T}} \leq 1$) to perform a direct and simplified showcase during the tracking control process.

\begin{figure}[h]
\begin{center}
        \includegraphics[scale=0.27]{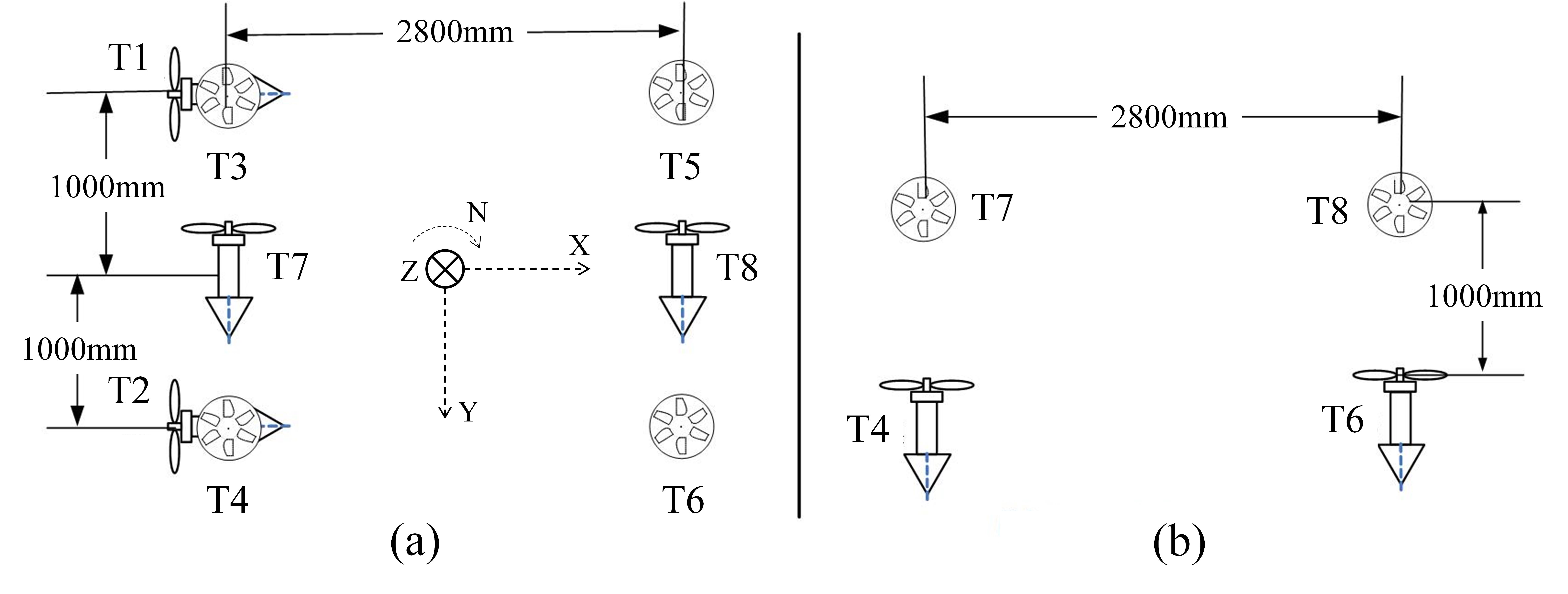}                         
        \caption*{Fig. 2. The thruster distribution in the "Yulong" UV propulsion system. (a) The top view, (b) The lateral view.}	
\end{center} 
\end{figure}

\subsection{Problem Statement}
In this subsection, the fault-tolerant control problem on the thruster system of the "Yulong" vehicle is modeled and explained. Requirements and constraints during the control process are introduced.

\subsubsection{Fault-tolerant Problem in UV Trajectory Tracking}
To achieve the accuracy and efficient control of the UV, the fault cases of the thruster system should be taken into consideration, where one or more of the thrusters might get broken and corresponding controls are proposed to sustain the movement and posture of the vehicle as desired, by deriving the normalized desired torques $\overline{\boldsymbol{\uptau}}_d$. The control of the torque outputs $\overline{\boldsymbol{\uptau}}$ is realized by the combined action of the thruster forces under fault conditions, which is allocated by approximation/optimization methods. The approximation/optimization deducts the optimal reallocation when the thrusters' working condition changes, and accomplishes the elimination of the torque errors during the process.

For UV with a multi-thruster system, the ideal fault-tolerant control is realized by,
\begin{equation}
||e||=||\overline{\boldsymbol{\uptau}}_d-\overline{\boldsymbol{\uptau}}||\,\rightarrow\,0\,,
\end{equation}
\begin{equation}
||\theta_e||=\arccos{\frac{\overline{\boldsymbol{\uptau}}_d \bullet \overline{\boldsymbol{\uptau}}}{||\overline{\boldsymbol{\uptau}}_d|| \cdot ||\overline{\boldsymbol{\uptau}}||}}\,\rightarrow\,0\,, 
\end{equation}
where $||e||$ is the magnitude error obtained by the vector norm of the difference between the desired and the actual torques of the vehicle, $||\theta_e||$ is the direction error that computes the arc cosine of the ratio between the multiplication of desired and actual torques and their vector norm. Besides the magnitude error, the direction error is also important to the trajectory tracking control of the underwater vehicle, as the vehicle’s movement is also determined by the direction displacement of its dynamic outputs (torques) along the axes. It is possible for the vehicle to have an acceptable magnitude dynamic error but meanwhile obtain excessive direction error, thus inducing non-ideal tracking results.

Hence, when designing the fault-tolerant control, the error values should be as small as possible to obtain good controlling results.

\subsubsection{Constraints on Fault-tolerant Control of the UV}
In the actual application, the controlling effect is always restricted by the physical constraints of the vehicle. The UV cannot provide infinite driving inputs such as torques/forces to complete the navigation, thus resulting in the problem of actuator saturation. Therefore driving restrictions are always applied on the UV to achieve a reliable and controllable navigation process. The maximum forces of the vehicle thrusters, derived from the maximum torques that can be offered by the vehicle body, are the essential constraints of the vehicle's controlling problem. 

To assess the influence of the constraints, maximum torques $\boldsymbol{\uptau_m}$ are introduced in the simulation part of this paper (see Fig. 3). By the definition given in Eq. (6), normalized torques $\boldsymbol{\overline{\uptau}}$ and normalized forces $\mathbf{\overline{T}}$ are supposed to have the limits of -1 to 1. The two variables are used to quantify the effect of the constraints during the trajectory tracking process in the simulation part. 

\section{GOA-based Fault-tolerant Trajectory Tracking Control (GFTC) Design}
The basic control architecture of the system is illustrated in Fig. 3. The design of the control strategy consists of two parts: (1) a trajectory tracking component formed by an outer loop of auxiliary kinematic control based on the position errors of the UV and an inner loop of dynamic torque controller based on the velocity state vector; (2) a fault-tolerant control component that identifies the fault cases of the UV propulsion system and reallocates the thruster force outputs based on Grasshopper optimization algorithm to eliminate the effect brought by the fault cases. Additionally, maximum thruster force outputs $\mathbf{T}_m$ are applied (see the bold frame in Fig. 3). Specially for the simulation of underwater environmental perturbations, current disturbance on torque outputs is considered. Moreover, the environmental noise at the step of forming vehicle positions is also involved. Details of the control design will be presented in this section.
\begin{figure*}[h]
\begin{center}
        \includegraphics[width=0.99\textwidth]{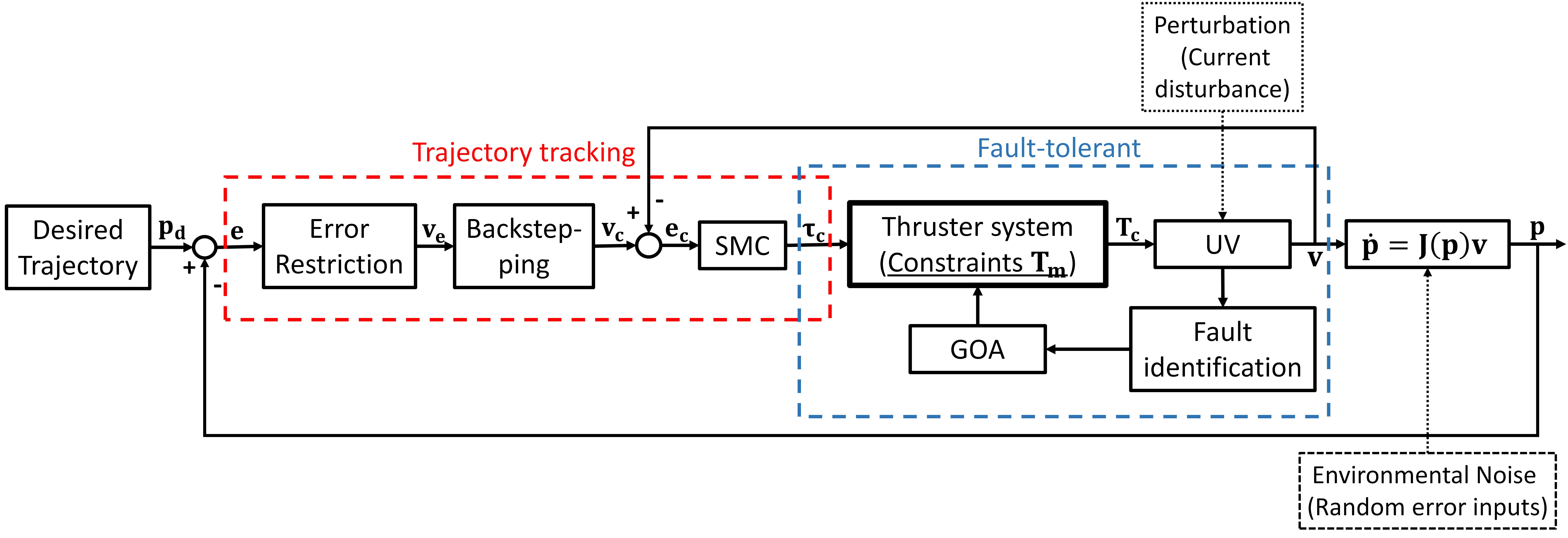}
        \caption*{Fig. 3. Schematic of the proposed fault-tolerant trajectory tracking control designed for the UV.}
\end{center} 
\end{figure*}

\subsection{Component of Trajectory Tracking Control}
In this section, the methods that form the kinematic and dynamic control of the UV are explained, with detailed equations and stability analysis given.
\subsubsection{Error Restricted Backstepping Control}
Suppose the maximum velocity vector of UV is
$\mathbf{v_m}=[v_{xm}~~v_{ym}~~v_{zm}~~v_{\psi m}]^T$, the input $\mathbf{e}(t)$ is the error between the desired and actual trajectories. a transfer function is defined to process the input trajectory errors within an acceptable range,
\begin{equation}
\mathbf{v}_e=f(\mathbf{e})\mu \mathbf{v}_m=\frac{\mathbf{e}(t)}{|\mathbf{e}(t)|+1}\mu \mathbf{v}_m\,,
\end{equation}
where $\mathbf{v_m}$ represents the maximum velocities of the UV DOFs at their corresponding axes; $\mu$ is a positive constant chosen according to the variation requirement and is assigned with 0.5 in this study.

Therefore, when $\mathbf{e}(t)\rightarrow0$, $\mathbf{v_e}=f(\mathbf{e}) \mu \mathbf{v_m}\rightarrow0$; and when $\mathbf{e}(t)\rightarrow\infty$, $\mathbf{v_e}=f(\mathbf{e}) \mu \mathbf{v_m}\rightarrow$ the restricted maximum velocity $\mu \mathbf{v_m}$. Based on the restriction, the transfer function $\mathbf{v}_e$ limits its control output in a smaller domain and meanwhile provides faster convergence to the input errors.

The processed outputs $\mathbf{v_e}$ after the convergence shall not exceed the maximum possible value of the UV velocity, and the definition of $f(\mathbf{e})$ provides a smooth transition of the UV at the beginning. Hence the speed-jump problem of the UV can be alleviated. The vector $\mathbf{v_e}$ at the four DOFs can be written as $\mathbf{v_e}=[v_{ex}~~v_{ey}~~v_{ez}~~v_{en}]^T$. Additionally, in the backstepping method, control functions for each subsystem are designed based on the Lyapunov techniques and generated to form the complete control law \cite{r28}. Therefore, based on Eq. (1) and the definition of the backstepping method, the error variables in the control law of the backstepping method are replaced by the restricted outputs processed in Eq. (10), the control law of the backstepping control can be derived as
\begin{gather}
    \mathbf{v_c}=
\begin{bmatrix}
   u_c\\
   v_c\\
   w_c\\
   r_c
\end{bmatrix}\\ \notag
=\begin{bmatrix}
   k(v_{ex}\cos{\psi}+v_{ey}\sin{\psi})+u_d\cos{v_{e\psi}}-v_d\sin{v_{e\psi}}\\
    k(-v_{ex}\sin{\psi}+v_{ey}\cos{\psi})+u_d\sin{v_{e\psi}}-v_d\cos{v_{e\psi}}\\
    w_d+k_zv_{ez}\\
    r_d+k_\psi v_{e\psi} 
\end{bmatrix}.
\end{gather}
where $k$, $k_z$ and $k_\psi$ are positive constants.

Then the processed control velocities $\mathbf{v_c}$ are passed to the UV, where they are calculated to keep pace with the desired trajectory through the dynamic model of the UV. Additionally, the stability of the refined backstepping control can be proved by constructing a Lyapunov function $\Gamma_0=\frac{1}{2}(e_x^2+e_y^2+e_z^2+e_\psi^2)$, whose derivative is less than and equal to zero (see Appendix A).

\subsubsection{Sliding Mode Control}
To design the sliding mode control, the desired dynamics (s) should be introduced. Based on Eq. (2) where the UV dynamic system is of the second order for the velocity $\mathbf{v}$, the dynamics can be designed as
\begin{equation}
\mathbf{s}=\left[\frac{d}{dt}+\lambda\right]^2\,\int \mathbf{e_v} dt=\mathbf{\dot{e}_v}+2\lambda \mathbf{e_v}+\lambda^{2}\int \mathbf{e_v} dt,
\end{equation}
where $\frac{d}{dt}$ is the derivative operator; $\mathbf{e_v}$ represents the errors given by the control velocities (see Fig. 3), $\mathbf{e_v}=\mathbf{v_c}-\mathbf{v}$; and $\lambda>0$ is a positive parameter \cite{r29}.\\
\indent Then take the derivative of $\mathbf{s}$, we can get
\begin{equation}
\mathbf{\dot{s}}=\mathbf{\ddot{e}_v}+2\lambda \mathbf{\dot{e}_v}+\lambda^{2} \mathbf{e_v}\,,
\end{equation}
where $\mathbf{\dot{e}_v}=\mathbf{\dot{v}_c}-\mathbf{\dot{v}}$

To keep the system states consistent with the desired dynamics, Eq. (13) should be equal to zero. This means the system states are on the sliding surface of the perfect tracking. At the same time, plug in the equation of the UV dynamic model (Eq. (2)),
\begin{eqnarray}
&&\;\;\;\mathbf{\dot{s}}=\mathbf{\ddot{e}_v}+2\lambda \mathbf{\dot{e}_v}+\lambda^{2} \mathbf{e_v}=0 \notag\\
&& \mathbf{\ddot{e}_v}+2\lambda (\mathbf{\dot{v}_c}-\mathbf{\dot{v}})+\lambda^{2} \mathbf{e_v}=0 \notag\\
&& \mathbf{\ddot{e}_v}+2\lambda (\mathbf{\dot{v}_c}-(\boldsymbol{\uptau}-\mathbf{C}\mathbf{v}-\mathbf{D}\mathbf{v}-\mathbf{g})\mathbf{M}^{-1})+\lambda^{2} \mathbf{e_v}=0 \notag\\
&& \boldsymbol{\uptau}=\mathbf{M}(\mathbf{\dot{v}_c}+\frac{\mathbf{\ddot{e}_v}}{2\lambda}+\frac{\lambda}{2}\mathbf{e_v})+\mathbf{C}\mathbf{v}+\mathbf{D}\mathbf{v}+\mathbf{g}\,.
\end{eqnarray}

The standard sliding mode control law is defined as
\begin{equation}
    \boldsymbol{\uptau}=\boldsymbol{\hat{\uptau}}+\boldsymbol{\uptau_c}\,,
\end{equation}
where $\boldsymbol{\hat{\uptau}}$ represents the major control law, which is continuous and model-based. It is designed to maintain the trajectory consistently on the sliding surface. $\boldsymbol{\uptau_c}$ represents the switching control law, dealing with the model uncertainty. When the trajectory is getting out of control, $\boldsymbol{\uptau_c}$ is used to push the trajectory back to the sliding surface and continue satisfactory tracking. For Eq. (14), supposing a simplification $\mathbf{\ddot{e}_v} \approx -k \mathbf{\dot{e}_v}$ based on the error acceleration feedback control to reduce computation complexity, the estimated item in the major control law
$\boldsymbol{\hat{\uptau}}$ can be deducted as
\begin{equation}
\boldsymbol{\hat{\uptau}}=\mathbf{\hat{M}}(\mathbf{\dot{v}_c}+\frac{-k \mathbf{\dot{e}_v}}{2\lambda}+\frac{\lambda}{2}\mathbf{e_v})+\mathbf{\hat{C}}\mathbf{v}+\mathbf{\hat{D}}\mathbf{v}+\mathbf{\hat{g}}\,,
\end{equation}
where $\hat{\mathbf{M}}$,  $\hat{\mathbf{C}}$,  $\hat{\mathbf{D}}$,  $\hat{\mathbf{g}}$ are the estimated values of $\mathbf{M}$, $\mathbf{C}$, $\mathbf{D}$ and $\mathbf{g}$; approximate values can be obtained from the practical case respectively \cite{r25}.\\
\indent The switching item $\boldsymbol{\uptau_c}$ in sliding mode control can be defined as
\begin{equation}
\boldsymbol{\uptau_c}=-\mathbf{K}_1\mathbf{s}-\mathbf{K}_2|s|^{r}sign(\mathbf{s})\,,
\end{equation}
where $sign(\mathbf{s})$ is the nonlinear sign function of $\mathbf{s}$; $\mathbf{K}_1$ and $\mathbf{K}_2$ are positive coefficients, $\mathbf{K}_1\geq \eta +F$ and $\mathbf{K}_2\geq \eta +F$, $\eta$ is the design parameter which is always chosen as a positive constant; $0<r< 1$; and $F$ represents the upper bound of the difference between the system actual output and the estimation,
\begin{equation}
F=|f(\mathbf{v})-\hat{f}(\mathbf{v})|.
\end{equation}

Additionally, an adaptive variation term $\widetilde{\mathbf{\uptau}}_{est}$ is added to the control law, where $\dot{\widetilde{\mathbf{\uptau}}}_{est}=\Gamma\mathbf{s}$ and $\Gamma$ represents a positive constant. Hence the final sliding mode control law is defined as
\begin{gather}
    \boldsymbol{\uptau}=\boldsymbol{\hat{\uptau}}+\widetilde{\mathbf{\uptau}}_{est}+\mathbf{\uptau_c} \notag\\
    =\mathbf{\hat{M}}(\mathbf{\dot{v}_c}+\frac{-k \mathbf{\dot{e}_v}}{2\lambda}+\frac{\lambda}{2}\mathbf{e_v})+\mathbf{\hat{C}}\mathbf{v}+\mathbf{\hat{D}}\mathbf{v}+\mathbf{\hat{g}} \notag \\ +\widetilde{\mathbf{\uptau}}_{est}-\mathbf{K}_1\mathbf{s}-\mathbf{K}_2|s|^{r}\rm sign(\mathbf{s})\,.
\end{gather}

Detailed proof of the SMC stability can be found in Appendix B.

\subsection{Component of Fault-tolerant Control}
The fault-tolerant control design is mainly built on the adjustment of the forces required by the thruster system, where the forces operate together and provide the torques as desired. The adjustment of the thruster forces is deducted by the grasshopper optimization algorithm (GOA), which efficiently eliminates the errors brought by the fault of the propulsion system after fault identification. Details of the control strategy are presented in this section.

\subsubsection{Weighting Matrix}
To quantify the degree of damage in the fault cases for the multi-thruster system, a weighting matrix $\mathbf{W}$ is introduced. The matrix $\mathbf{W}$ decides the service condition of the thruster, which is usually defined as a diagonal matrix,
\begin{equation}
    \mathbf{W}=
    \begin{bmatrix}
       w_1 & 0 & 0 & 0 & 0 & 0 & 0 & 0\\
       0 & w_2 & 0 & 0 & 0 & 0 & 0 & 0\\
       0 & 0 & w_3 & 0 & 0 & 0 & 0 & 0\\
       0 & 0 & 0 & w_4 & 0 & 0 & 0 & 0\\
       0 & 0 & 0 & 0 & w_5 & 0 & 0 & 0\\
       0 & 0 & 0 & 0 & 0 & w_6 & 0 & 0\\
       0 & 0 & 0 & 0 & 0 & 0 & w_7 & 0\\
       0 & 0 & 0 & 0 & 0 & 0 & 0 & w_8
    \end{bmatrix}
\end{equation}
where $w_j>0$ is the weight of the $j^{th}$ thruster. If all the thrusters are working in the desired condition with no power loss, $\mathbf{W}$ will be a unit matrix, meaning all $w_j=1$. If there is power loss for any of the thrusters, its corresponding weight will be reduced by the degree of the loss. For example, when $T_1$ thruster attains 20\% of power loss, $w_2$ in the weighting matrix is assigned as 0.8 respectively.

As the relation between the thruster forces and the vehicle torques at different states are defined and given in Eq. (7), the following transition between the torques and forces in the fault cases is defined as,
\begin{equation}
\overline{\boldsymbol{\uptau}}
=\overline{\mathbf{B}}\mathbf{W}\mathbf{\overline{T}},
\end{equation}
where $\mathbf{\overline{T}}$ is the control parameters in the UV case, which is deducted by the optimization method, such as the GOA method used in this study.

Additionally, as a comparison to the GOA method, the weighted pseudo-inverse matrix method is used, which is determined based on the defined weighting matrix,
\begin{equation}
    \mathbf{\overline{T}}=\overline{\mathbf{B}}_{w}^{+}\, \overline{\boldsymbol{\uptau}}_{d}
    =(\mathbf{W}\overline{\mathbf{B}}^{T}(\overline{\mathbf{B}}\mathbf{W}\overline{\mathbf{B}}^{T})^{-1})\overline{\boldsymbol{\uptau}}_{d},
\end{equation}
where $\overline{\mathbf{B}}_{w}^{+}$ is the matrix that transmits the damage information to the propulsion system and meanwhile make the adjustment accordingly. Thus, the thruster force results under fault cases can be deducted. For example, if the thruster $T_1$ can only provide 70\% of power after encountering a power loss of 30\%, the weighted pseudo-inverse matrix method will request larger output ($\frac{1}{0.7}\times$ original force) from $T_1$ such that the same force can be achieved after weakened by 30\% of power loss.

Then T-approximation or S-approximation methods are applied for achieving force results within the range of the thruster force maximum, which is generally denoted as pseudo-inverse (P-I) matrix approximation. T-approximation restricts all normalized forces $\mathbf{\overline{T}}$ between [-1, 1] by subtracting/adding the excessive part of the states whose value is larger than 1 or smaller than -1, where
\begin{equation}
\begin{aligned}
&\overline{\mathbf{T}}_t=    \left\{
            \begin{array}{lr}  
            \overline{T}_i, &  \overline{T}_i \in [-1,1]\\  
            1, &  \overline{T}_i > 1\\  
            -1, &  \overline{T}_i < -1\\
             \end{array}
    \right.\\
    \notag
\end{aligned}
\end{equation}

S-approximation realizes the limits of [-1, 1] by timing the reciprocal ratio of the largest normalized force for all states, where
\begin{equation}
  \overline{\mathbf{T}}_s=\frac{1}{max(\overline{T}_i)}\overline{\mathbf{T}}, i=1,2,...,8.
  \notag
\end{equation}
For example, in S-approximation, if the largest normalized force for one of the states reaches 2, all normalized forces will be multiplied by the ratio of $\frac{1}{2}$ to guarantee they do not exceed the limits of -1 to 1.

In the simulation section of this study, T-approximation method is used as the typical pseudo-inverse (P-I) matrix approximation to work as a comparison of the proposed GOA-based FTC. The T-approximation has wider application in practical cases of the underwater vehicle FTC due to it generally produces smaller errors compared to the S-approximation \cite{Liu2011,danjie2021OE}.

\subsubsection{Grasshopper Optimization Algorithm}
The grasshopper optimization algorithm is newly raised in 2017 \cite{Saremi2017}. As a developed algorithm based on the theory of swarm intelligence that imitates the activity of grasshoppers, GOA shows better performance than the traditional swarm intelligence algorithms due to that it finds a satisfactory balance between fast speed of convergence and accuracy based on its form switch between “adults” and “larvae”. The fast convergence is realized when GOA searches globally based on the position of each agent under its "adult" form, which explores on a large scale in an attractive manner among the agents; while the accuracy is achieved by shrinking the range and keeping a repulsive zone based on the best agent under "larvae" form, which avoids the local minimum.

According to the movement of the grasshopper groups, a mathematical model can be defined to describe their swarming behavior \cite{swarming1,swarming2}
\begin{equation}
    X_{i}=\sum_{j=1,j\neq i}^Ns(|x_j-x_i|)\frac{x_j-x_i}{d_{ij}}-G_i+A_i,
\end{equation}
where $X_{i}$ represents the next position of the $i_{\rm th}$ grasshopper; $s(r)$ is the social interaction function where it is optimized as $s(r)=0.5e^{-r/1.5}-e^{-r}$. The item $|x_j-x_i|$ is the distance between the current position of the $i_{\rm th}$ and $j_{\rm th}$ grasshopper, $(x_j-x_i)/d_{ij}$ is the unit vector pointing from the position of the $i_{\rm th}$ grasshopper to the $j_{\rm th}$ grasshopper. $G_i$ represents the Gravity force at the $i_{\rm th}$ grasshopper; $A_i$ is the wind advection that is assumed to be always towards the target, $T_i$.

Based on the assumptions made in this control case, where the gravity force is neglected and the wind force is always towards the target, Eq. (23) can be converted into
\begin{equation}
    X_{i}^{d}=c(\sum_{j=1,j\neq i}^N c\frac{ub_d-lb_d}{2} s(|x_j-x_i|)\frac{x_j-x_i}{d_{ij}})+T_d
\end{equation}
where $c$ is a decreasing coefficient that shrinks the comfort zone, repulsion zone and attraction zone, which is determined as $c=c_{\rm max}-l(c_{\rm max}-c_{\rm min})/L$, $c_{\rm max}$ is the maximum value, $c_{\rm min}$ is the minimum value, $l$ indicates the current iteration, and $L$ is the maximum number of iterations. In this work, we assign $c_{\rm max}=1$ and $c_{\rm min}=0.00001$ by trial and error. The variable $ub_d$ represents the upper bound of the case while the $lb_d$ represents the lower bound, which are 1 and -1 in this design. $T_d$ is the desired solution of the current iteration. These parameters are used to attain the fast convergence of the optimization, by increasing the speed of updating the local solution in relation to the increment of the iteration times, thus leading to the efficient searching result of the GOA method \cite{Meraihi2021}.

The pseudocode of the GOA applied on the UV thruster forces reallocation can be concluded as follows (see Algorithm 1), with the fitness evaluation substituted by the error evaluation, given in Eq.s (8) and (9).
\begin{figure}[h]
\begin{center}
        \includegraphics[scale=0.4]{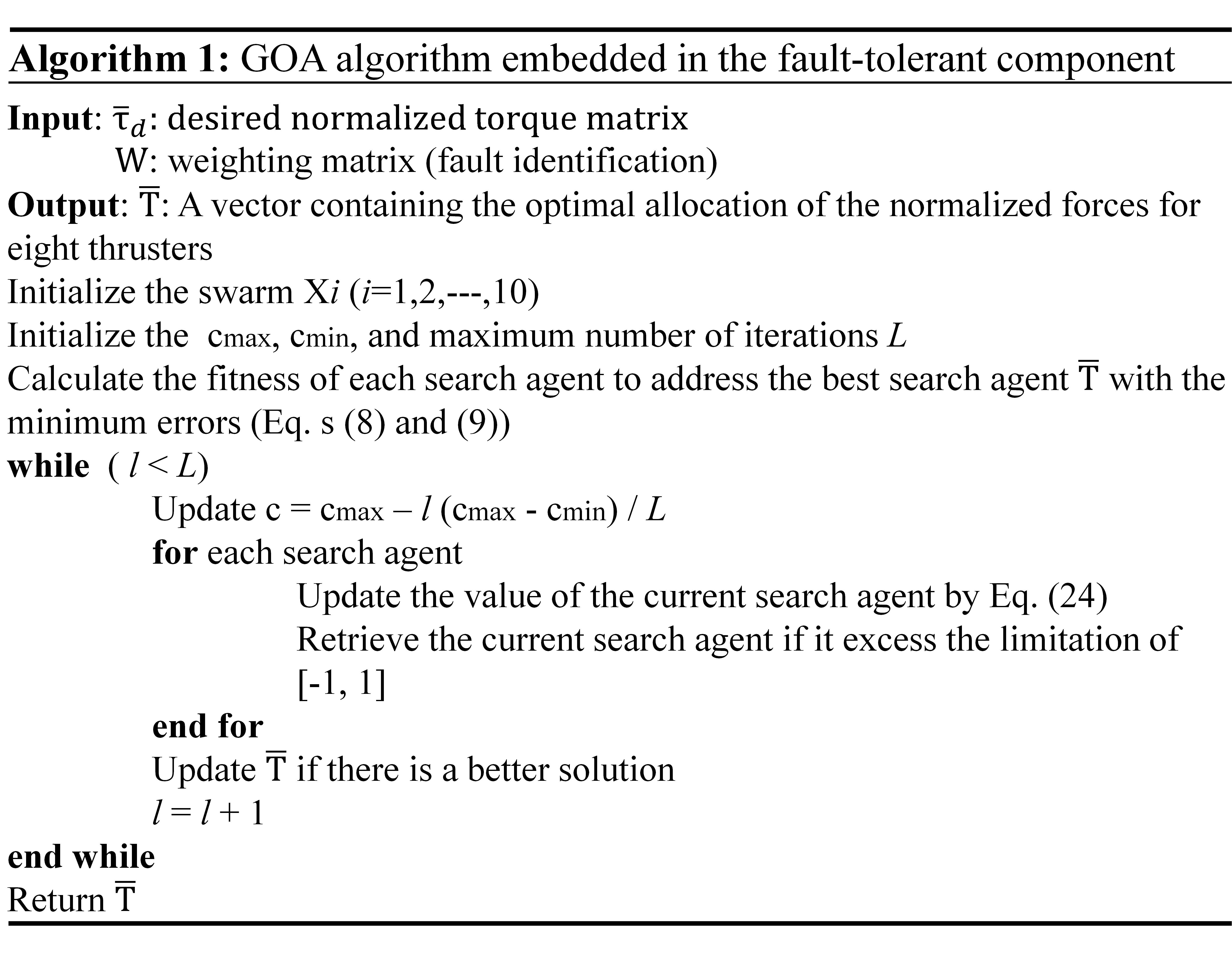} 
\end{center} 
\end{figure}

The flow of the proposed GOA method can be concluded as follows:

\noindent 1) First initialize the swarm, with 10 groups of eight random numbers between -1 and 1 representing the normalized eight-thruster group and each group is regarded as a search agent;

\noindent 2) Next calculate the fitness of each search agent and address the agent with the minimum errors as the best based on the objective function combined by Eq. (8) and (9), which is $||e||+||\theta_{e}||\,\rightarrow\,0\,$, the constraints for each agent are set between [-1, 1];

\noindent 3) Update the parameter c according to the iteration time to accelerate the convergence. If iteration time reaches maximum then stop, otherwise continue the update of c;

\noindent 4) Update positions (values) of search agents based on Eq. (24) and compare the fitness of the updated agents with the agent of the best fitness. If the updated fitness turns out to be better, updates the position of the agents, otherwise do not update.

\noindent 5) Update the iteration time, and repeat the loop from step 3).

\subsection{Component of Perturbations}
As the "Yulong" UV model is designed for dam detection, which usually operates at the shoreside underwater condition, the perturbation of currents can be considered in a regular combination form of wave functions as \cite{wave1,wave2,wave3}
\begin{equation}
    {\tau_{p}}=
      A_{p1}\cos(\omega_{p1}t)\sin(\omega_{p2}t)+A_{p2}\cos{(\omega_{p3}t)}\sin{(\omega_{p4}t)},
\end{equation}
where $A_{p1}$, $A_{p2}$ and $\omega_{p1}$ to $\omega_{p4}$ are random coefficients, which are chosen to synthesize the randomness of the currents to appropriately address the underwater environmental perturbation. $A_{p1}$ and $A_{p2}$ are assigned within the range of $10\%$ of the torque outputs at four axes, for example, if $\tau_{x}$ output is about $100N$, the assignment range will be $[-10, 10]$. Coefficients $\omega_{p1}$ to $\omega_{p4}$ are chosen with the range of -1 to 1.

In addition, considering the effect of environmental noise that produces perturbation to the data transfer at the stage of forming positions, an random error input is given in the simulation. The random error is supposed to be within [-0.1, 0.1] and filtered by sensors, which corresponds to the practical UV case.

\section{SIMULATION RESULTS AND ANALYSIS }
In this section, the polygonal line and helix trajectory tracking simulation results of the proposed FTC and conventional approximation methods are presented and analyzed. Fault cases of single-fault (one thruster broken) and double-fault (two thrusters broken) are applied due to their frequent occurrence.

\subsection{Helix Tracking}
In this section, one of the thrusters $T_1$ is supposed to be broken, where 100\% of power is lost. The initial position of the desired helix trajectory is set at $(0,0,0,0)$, while the initial position of the control trajectories is set at $(0,10,0,0)$. The difference of the initial positions is given to test the correction ability of the two tracking strategies when they start with a certain amount of deviation at one of the axes, i.e. the y axis. Assuming the desired trajectory is given as $x_d=10\sin{0.2t}$, $y_d=10-10\cos{0.2t}$, $z_d=0.5t$ and $\psi_d=0.2t$ with the simulation time continuing for 50 seconds. 

The GFTC (in red dash) tracking result quickly eliminates the initial error and follows the desired helix trajectory till the end of the simulation yet the $T_1$ thruster is supposed to be completely broken (Fig. 4(a)). The P-I approximation (in blue) cannot coincide with the desired helix under the single-fault case, where the deviation of abrupt variations is created at the beginning and trajectory distortions are produced throughout the whole process. Therefore when the dynamic constraints are considered, the P-I method fails to compensate for the power loss of a single thruster, which induces increasing errors and excessive velocities with abrupt jumps given in Fig.s 4(b) and (c); while the GFTC achieves smooth error and control velocity curves that indicate the satisfactory tracking performance of the method. Moreover, the "GFTC-P" (in pink) results consider the effect of perturbations brought by the currents and environmental noise when addressing the position information for the vehicle. The GFTC-P result under the effect of perturbations sustains a similar tracking trajectory with the unaffected GFTC, which verifies the robustness of the proposed control. 

\begin{figure}[t]
\begin{center}
        \includegraphics[scale=0.48]{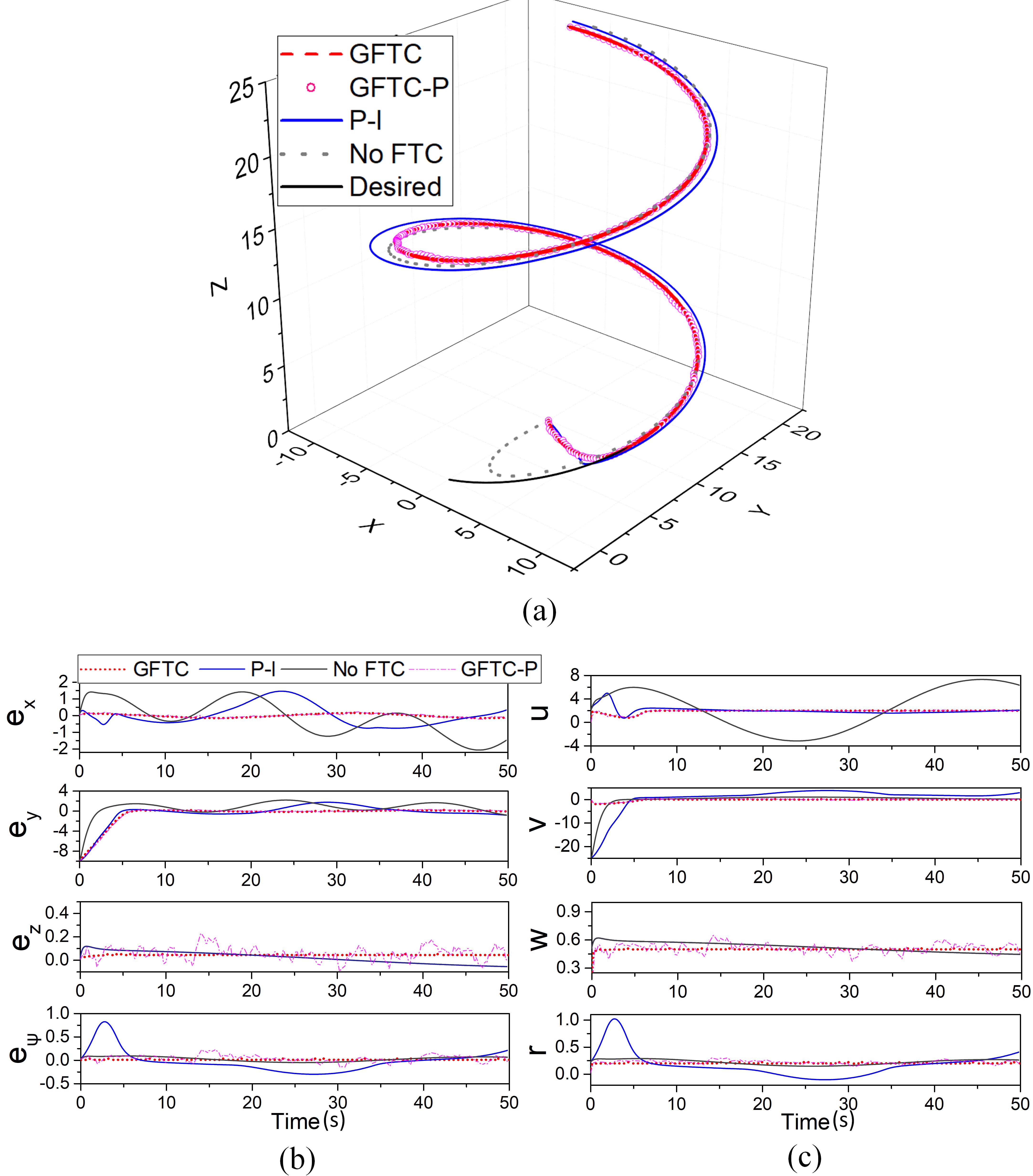} \captionsetup{justification=centering}
        \caption*{Fig. 4. Helix tracking results using the GFTC with or without perturbations and the P-I approximation-based FTC under the single-fault case. (a) Comparison of trajectories, (b) Comparison of tracking errors, (c) Comparison of control velocities.}
\end{center} 
\end{figure} 

\begin{table}[h]
\centering
\captionsetup{justification=centering}
\caption*{TABLE \uppercase\expandafter{\romannumeral1}. Maximum velocities of the GFTC with or without perturbations and the P-I based FTC under the single-fault case when tracking the helix}
\begin{tabular}{|c|c|c|c|c|}
\hline
& $u_c$ (m/s) & $v_c$ (m/s) & $w_c$ (m/s) & $r_c$ (m/s)\\
\hline
GFTC & 2.0008 & -1.8978 & 0.5002 & 0.2016\\
\hline
P-I & 4.9719 & -25 & 0.6125 & 1.0034\\
\hline
GFTC-P & 2.0150 & -1.8944 & 0.6404 & 0.2836\\
\hline
\end{tabular}
\end{table}

Tracking errors of the GFTC method (in red), P-I approximation-based FTC (in blue), single-fault case without FTC (in grey) and GFTC-P with the effect of perturbations (in pink) are given in Fig. 4(b). The error curve of the GFTC method eliminates the initial deviation and quickly converges to zero. While the error of the P-I approximation-based FTC presents obvious fluctuations and cannot be eliminated in all axes, furthermore, the P-I error curve attains even larger fluctuations compared to the case without FTC at the $\psi$ axis, which supports the trajectory tracking performance given in Fig. 4(a). This shows the failure of P-I approximation FTC on keeping the desired helix tracking in the single-fault condition when dynamic constraints are applied, yet the GFTC method accomplishes the fault-tolerant trajectory tracking task with satisfactory kinematic outputs. This conclusion is also supported by the velocity variations in Fig. 4(c). The P-I method deducts largely excessive speeds at the x and y axes, where the maximum of 4.9719 m/s and -25 m/s are required, but the GFTC method satisfactorily restricts the velocity at x and y axes within the constraints of [-2, 2]m/s, with the maximum outputs at 2.0008 m/s and -1.8978 m/s (TABLE \uppercase\expandafter{\romannumeral1}). The GFTC also achieves much smoother velocity curves compared to the P-I method for all axes. Even when considering the perturbation effect in GFTC-P simulation, though small chattering is performed, the errors as well as the control velocities are successfully restricted in an acceptable range, with the maximum of 2.015m/s and -1.8944m/s at the x and y axes, and far less requirement of control velocity at the $\psi$ axis. Hence the effectiveness of the proposed GFTC method in tracking the desired trajectory under the single-fault case is verified even when external perturbations are given. 
\subsection{3D Polygonal Line Tracking} 
A 3D polygonal line is applied in this section as the reference tracking trajectory, as the "YuLong" UV usually navigates in a movement similar to the polygonal line to detect the dam damage. 

The initial position of the desired trajectory is set at $(0,0,0,0)$, while the initial position of the control trajectories is set at $(0,2.5,0,0)$. A specific polygonal line function is applied and the simulation continues for 20 seconds:  
\begin{equation}
\begin{aligned}
&x_d= t, 0 \leq t \leq 20,\\
&y_d=    \left\{
            \begin{array}{lr}  
            t, &  0 \leq t \leq 5\\  
            5, & 5 < t\leq 10\\  
            t-5, & 10 < t\leq 15\\
            10, & 15 < t\leq 20 
             \end{array}
    \right.\\
&z_d=    \left\{
            \begin{array}{lr}  
            t, &  0 \leq t \leq 5\\  
            5, & 5 < t\leq 10\\  
            t-5, & 10 < t\leq 15\\
            10, & 15 < t\leq 20 
             \end{array}
   \right.\\
&\psi_d= 0.2, 0 \leq t \leq 20.\\
    \notag
\end{aligned}
\end{equation}

\subsubsection{Single-fault Case}
One of the thrusters $T_8$ is supposed to be broken, with 100\% of power lost. The tracking trajectory results are shown in Fig. 5(a). The GFTC (in red dash) has retained the polygonal line trajectory as desired after eliminating the initial error at the y axis, neglecting the power loss of the thruster. Moreover, the GFTC-P (in pink) results which consider the effect of perturbations sustain a similar tracking trajectory with the unaffected condition, which verifies the robustness of the proposed tracking control. The P-I method (in blue) fails to catch up with the desired trajectory especially at the turning point where larger dynamic inputs are needed. The P-I method cannot make up for the loss of the propulsive force when physical constraints (torque/force maximum) are involved, thus producing errors with large fluctuations as well as excessive control velocities presented in Fig.s 5(b) and (c). 

\begin{figure}[t]
\begin{center}
        \includegraphics[scale=0.47]{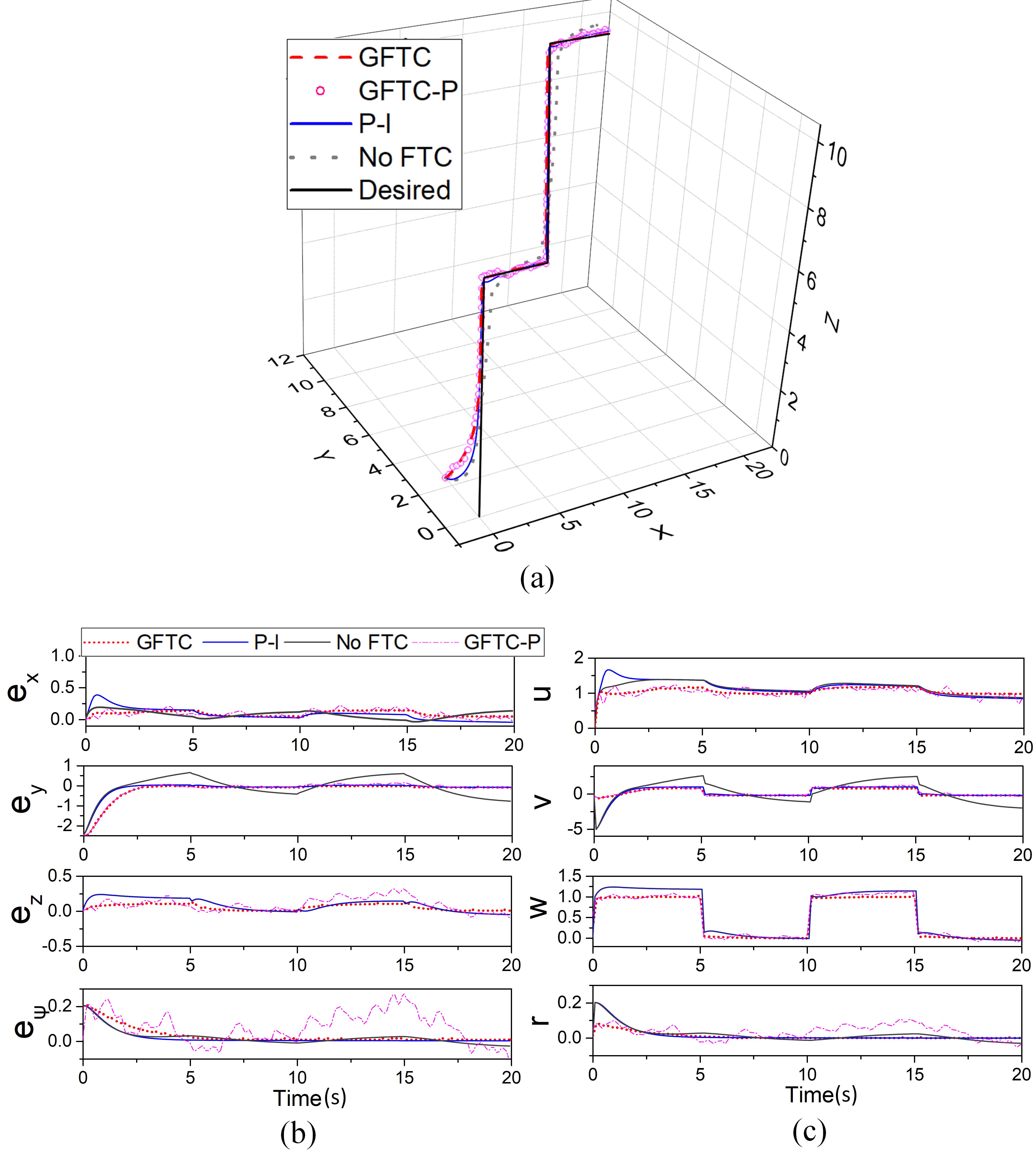}                         \captionsetup{justification=centering}
        \caption*{Fig. 5. Polygonal line tracking results using the GFTC with or without perturbations and the P-I approximation-based FTC under the single-fault case. (a) Comparison of trajectories, (b) Comparison of tracking errors, (c) Comparison of control velocities.}
\end{center} 
\end{figure} 

\begin{table}[h]
\centering
\captionsetup{justification=centering}
\caption*{TABLE \uppercase\expandafter{\romannumeral2}. Maximum velocities under the single-fault case when tracking the polygonal Line}
\begin{tabular}{|c|c|c|c|c|}
\hline
& $u_c$ (m/s) & $v_c$ (m/s) & $w_c$ (m/s) & $r_c$ (m/s)\\
\hline
GFTC & 1.1738 & 0.8212 & 1.0012 & 0.0782\\
\hline
P-I & 1.7278 & -5.0713 & 1.2289 & 0.2\\
\hline
GFTC-P & 1.2410 & 1.2188 & 1.1192 & 0.1035\\
\hline
\end{tabular}
\end{table}

The errors at four axes are presented in Fig. 5(b), where the GFTC (in red) successfully eliminates the tracking errors. While for the P-I approximation-based FTC (in blue), its result fails to eliminate the error once the sharp turning is required by the trajectory, as the excessive dynamic outputs deducted by the P-I method cannot be satisfied when the dynamic constraints are applied, thus inducing the large trajectory deviation at the turning section. Velocity variations at four axes are shown in Fig. 5(c), the P-I method performs a sharp fluctuation and fails to retain the control velocity within the desired range at the y axis (see y axis in TABLE \uppercase\expandafter{\romannumeral2}). The velocity at the y axis of P-I method reaches a dramatic value of -5.0713 m/s, largely exceeding the desired range of the vehicle that is preset at -2m/s to 2m/s. At the same time, the GFTC successfully limits the kinematic outputs within the constraints and presents a smooth curve of much smaller fluctuations. In addition, when considering the perturbations in the GFTC-P simulation (in pink), though chattering issues are presented, errors are constrained within an acceptable range and kinematic outputs perform a smaller range compared to the P-I method at most axes, with the maximum of 1.2410m/s and 1.2188m/s at the x and y axes. These results indicate the effectiveness and robustness of the proposed GOA-based FTC. 

\subsubsection{Double-fault Case}
In this section, the effect of the GFTC, P-I approximation based FTC and GFTC considering environmental perturbations are compared, supposing two thrusters ($T_1$ and $T_8$) of the propulsion system encounter power loss of 100\%.

The GFTC method (in red dash) successfully eliminates the initial errors at the y axis and retains the tracking trajectory as desired till the end (Fig. 6(a)). The GFTC-P (in pink) results under the effect of perturbations sustain a similar tracking trajectory with the unaffected GFTC trajectory, and at the second turning section it performs more smooth tracking curves than the first one, which verifies its robustness for tracking the desired polygonal line even under the double-fault case. At the same time, the P-I approximation based FTC fails to track the desired trajectory and even presents a much larger deviation compared to its single-fault case (Fig. 5(a)). This demonstrates that the GFTC method is capable of balancing off the tracking errors whenever the damage degree of power loss in the thruster system differs, thus proving the robustness of the proposed FTC.

Similarly, as under the single-fault case, the error curve of GFTC method under double-fault case eliminates the initial deviation and quickly converges to zero (Fig. 6(b)). However, the P-I method presents fierce error vibrations in x, y and $\psi$ axes compared to the single-fault case, which is even worse than the performance of double-fault case without FTC. This shows that the P-I approximation-based FTC is heavily affected by the damage degree of the UV propulsion system and it cannot balance off the error produced by the excessive power loss of the thrusters, e.g. the double-fault case. This conclusion is also supported by the velocity variations in Fig. 6(c), where the P-I method cannot be limited within the allowable range due to the thruster power loss, with excessive maximum velocities arriving at 17.5815m/s for the x axis, 25.9814m/s for the y axis and 1.4485m/s for the $\psi$ axis given in TABLE \uppercase\expandafter{\romannumeral3}, resulting in the complete tracking failure shown in Fig. 6(a). The velocities of the GFTC method maintain within the allowable domain throughout the whole process, neglecting the change of fault cases. The GFTC-P simulations that involve perturbations in a practical underwater environment also perform successful restriction of the errors and the control velocities within the supposed range, with the maximum of 1.2473m/s and 0.9646m/s at the x and y axes. Therefore, the effectiveness of the proposed GFTC method in tracking a desired polygonal line is verified whenever single-fault or double-fault cases are applied.  

\begin{figure}[t]
\begin{center}
        \includegraphics[scale=0.47]{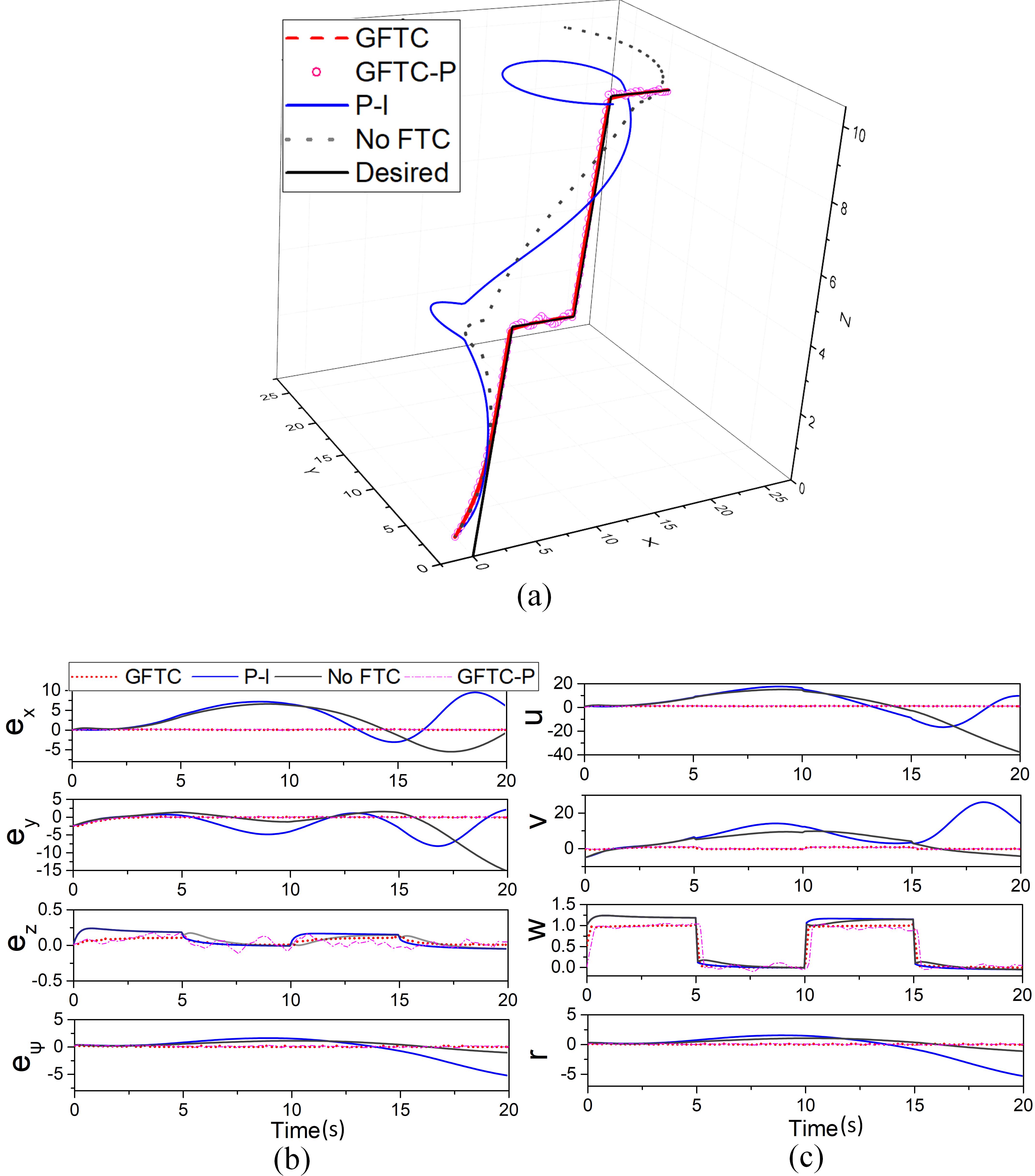}                         \captionsetup{justification=centering}
        \caption*{Fig. 6. Polygonal line tracking results using the GFTC with or without perturbations and the P-I approximation-based FTC under the double-fault case. (a) Comparison of trajectories, (b) Comparison of tracking errors, (c) Comparison of control velocities.}
\end{center} 
\end{figure} 

\begin{table}[h]
\centering
\captionsetup{justification=centering}
\caption*{TABLE \uppercase\expandafter{\romannumeral3}. Maximum velocities under the double-fault case when tracking the polygonal line}
\begin{tabular}{|c|c|c|c|c|}
\hline
& $u_c$ (m/s) & $v_c$ (m/s) & $w_c$ (m/s) & $r_c$ (m/s)\\
\hline
GFTC & 1.1710 & 0.8158 & 0.9989 & 0.0782\\
\hline
P-I & 17.5815 & 25.9814 & 1.2289 & 1.4485\\
\hline
GFTC-P & 1.2473 & 0.9646 & 1.0367 & -0.0885\\
\hline
\end{tabular}
\end{table}

\section{Conclusion}
In this paper, the fault-tolerant trajectory tracking problem for the "Yulong" UV is resolved by a Grasshopper Optimization and backstepping \& SMC-based cascade control (GFTC). The GFTC strategy applies a refined backstepping algorithm to restrict the kinematic outputs; and the Grasshopper Optimization Algorithm (GOA) is used to achieve optimized thruster force reallocation within the allowable domain. When encountering fault cases in tracking the polygonal line or helix, the trajectory tracking errors of the GFTC are largely alleviated and the actuator saturation problem is eliminated, compared to the traditional FTCs such as the weighted pseudo-inverse matrix approximation-based methods. In addition, the robustness of the proposed FTC is also verified when environmental perturbations are involved, which serves as the basis of the experimental study on practical applications that will be extended in the future.

\appendices
\section{Proof of the Error Restricted Backstepping Control Stability}
According to the Lyapunov stability theory, a special Lyapunov function $\Gamma_0$ is chosen,
\begin{equation}
\Gamma_0=\frac{1}{2}(e_x^2+e_y^2+e_z^2+e_\psi^2)\,.
\end{equation}

By Eq.s (1) and (11), the derivative of Eq. (25) can be obtained to prove the stability of the backstepping system,
\begin{equation}
    \begin{aligned}
\,\,\dot{\Gamma}_0=&e_x\,\dot{e}_x+e_y\,\dot{e}_y+e_z\,\dot{e}_z+e_\psi\,\dot{e}_\psi\\&
=e_x\,(\dot{x}_d-\dot{x})+e_y\,(\dot{y}_d-\dot{y})\\&\;\;\;
+e_z(\dot{z}_d-\dot{z})+e_\psi\,(\dot{\psi}_d-\dot{\psi})\\&
=e_x\,[(\cos{\psi_d}u_d-\sin{\psi_d}v_d)-(\cos{\psi} u_c-\sin{\psi} v_c)]\\&\;\;\;
+e_y\,[(\sin{\psi_d}u_d+\cos{\psi_d}v_d)-(\sin{\psi} u_c+\cos{\psi} v_c)]\\&\;\;\;
+e_z(w_d-w_c)+e_\psi\,(r_d-r_c)\\&
=e_x\,[(\cos{\psi_d}u_d-\sin{\psi_d}v_d)\\&\;\;\;
-(kv_{ex}+u_d(\cos{\psi}\cos{v_{e\psi}}-\sin{\psi}\sin{v_{e\psi}})\\&\;\;\;
+v_d(\sin{\psi}\cos{v_{e\psi}}-\cos{\psi}\sin{v_{e\psi}}))]\\&\;\;\;
+e_y\,[(\sin{\psi_d}u_d+\cos{\psi_d}v_d)\\&\;\;\;
-(kv_{ey}+u_d(\sin{\psi}\cos{v_{e\psi}}+\cos{\psi}\sin{v_{e\psi}})\\&\;\;\;
-v_d(\sin{\psi}\sin{v_{e\psi}}-\cos{\psi}\cos{v_{e\psi}}))]\\&\;\;\;
+e_z(-k_zv_{ez})+e_\psi\,(-k_\psi v_{e\psi})
\\&
\;\,\leq -ke_xv_{ex}-ke_yv_{ey}-k_ze_zv_{ez}-k_\psi e_\psi v_{e\psi}\,.
    \end{aligned}
\end{equation}

According to the definition of $\mathbf{v_{e}}$, $\mathbf{e}(t)$ are of the same sign (see definition of Eq. (10)); $k$, $k_z$, $k_\psi$ are positive constants. The result of Eq. (26) is believed to be less than and equal to zero, which demonstrates the stability of the designed refined backstepping controller.

\section{Proof of the SMC stability}
To prove the stability of the SMC, construct a Lyapunov function,
\begin{equation}
    \mathbf{V}=\frac{1}{4\lambda}\mathbf{s}^{T}\mathbf{M}\mathbf{s}+\frac{1}{2}\mathbf{Q}^{T}\Gamma^{-1}\mathbf{Q}\,,
\end{equation}
where $\mathbf{Q}=\widetilde{\mathbf{\uptau}}_r-\widetilde{\mathbf{\uptau}}_{est}$ and $\widetilde{\mathbf{\uptau}}_r=\mathbf{\widetilde{M}}\mathbf{\dot{v}}_r+\mathbf{\widetilde{C}}\mathbf{v}_r+\mathbf{\widetilde{D}}\mathbf{v}+\mathbf{\widetilde{g}}$.

Previously we have given $\mathbf{e_v}=\mathbf{v_c}-\mathbf{v}$, and $\mathbf{s}
\mathbf{\dot{e}_v}+2\lambda \mathbf{e_v}+\lambda^{2}\int \mathbf{e_v} dt$, such that two equations can be deducted as, 
\begin{equation}
    \mathbf{v}=\mathbf{v_c}-\frac{\mathbf{s}-\mathbf{\dot{e}_v}-\lambda^{2}\int \mathbf{e_v} dt}{2\lambda},
\end{equation}
\begin{equation}
    \mathbf{\dot{v}}=\mathbf{\dot{v}_c}-\frac{\mathbf{\dot{s}}-\mathbf{\ddot{e}_v}-\lambda^{2}\mathbf{e_v}}{2\lambda},
\end{equation}
therefore the following items can be defined,
\begin{equation}
    \mathbf{v_r}=\mathbf{v_c}+\frac{\mathbf{\dot{e}_v}+\lambda^{2}\int \mathbf{e_v} dt}{2\lambda},
\end{equation}
\begin{equation}
    \mathbf{\dot{v}_r}=\mathbf{\dot{v}_c}+\frac{\mathbf{\ddot{e}_v}+\lambda^{2}\mathbf{e_v}}{2\lambda}.
\end{equation}

By substituting into Eq. (2),
\begin{gather}
    \mathbf{M}\frac{\mathbf{\dot{s}}}{2\lambda}+\mathbf{C}\frac{\mathbf{s}}{2\lambda} \notag \\
    =\mathbf{M}(\mathbf{\dot{v}_c}+\frac{\mathbf{\ddot{e}_v}+\lambda^{2}\mathbf{e_v}}{2\lambda})+\mathbf{C}(\mathbf{v_c}+\frac{\mathbf{\dot{e}_v}+\lambda^{2}\int \mathbf{e_v} dt}{2\lambda}) \notag\\
    +\mathbf{D}\mathbf{v}+g-\mathbf{\uptau}
    =\mathbf{M}\mathbf{\dot{v}_r}+\mathbf{C}\mathbf{v_r}+\mathbf{D}\mathbf{v}+g-\mathbf{\uptau}.
\end{gather}

Based on previous definitions, the derivative of Eq. (27) can be simplified as,
\begin{eqnarray}
&&\mathbf{\dot{V}}=\frac{1}{4\lambda}(\mathbf{s}^{T}\mathbf{\dot{M}}\mathbf{s}+\mathbf{\dot{s}}^{T}\mathbf{M}\mathbf{s}+\mathbf{s}^{T}\mathbf{M}\mathbf{\dot{s}}) \notag \\
&&+\frac{1}{2}\mathbf{\dot{Q}}^{T}\Gamma^{-1}\mathbf{Q}+\frac{1}{2}\mathbf{Q}^{T}\Gamma^{-1}\mathbf{\dot{Q}} \notag\\
&& =\frac{1}{2\lambda}\mathbf{s}^{T}(\mathbf{M}\mathbf{\dot{s}}+\mathbf{C}\mathbf{s})+\frac{1}{2}\mathbf{\dot{Q}}^{T}\Gamma^{-1}\mathbf{Q}+\frac{1}{2}\mathbf{Q}^{T}\Gamma^{-1}\mathbf{\dot{Q}} \notag\\
&& = \mathbf{s}^{T}(\mathbf{M}\mathbf{\dot{v}_r}+\mathbf{C}\mathbf{v_r}+\mathbf{D}\mathbf{v}+g-\mathbf{\uptau})+\mathbf{\dot{Q}}^{T}\Gamma^{-1}\mathbf{Q}. 
\end{eqnarray}

By substituting Eq. (19),
\begin{gather}
\mathbf{\dot{V}}
= \mathbf{s}^{T}(\mathbf{M}\mathbf{\dot{v}_r}+\mathbf{C}\mathbf{v_r}+\mathbf{D}\mathbf{v}+g-\mathbf{\uptau}) \notag\\
+(\dot{\widetilde{\mathbf{\uptau}}}_r-\dot{\widetilde{\mathbf{\uptau}}}_{est})^{T}\Gamma^{-1}\mathbf{Q} \notag\\
=-\mathbf{s}^{T}(\mathbf{K}_1\mathbf{s}+\mathbf{K}_2|\mathbf{s}|^{r}sign(\mathbf{s}))+(\dot{\widetilde{\mathbf{\uptau}}}_r)^{T}\Gamma^{-1}\mathbf{Q}.  
\end{gather}

The dynamic item $\mathbf{\widetilde{\uptau}}_r$ is bounded due to the slow velocity of the underwater vehicle and $\mathbf{s}^{T}(\mathbf{K}_1\mathbf{s}+\mathbf{K}_2|\mathbf{s}|^{r}\rm{sign}(\mathbf{s}))\geq(\dot{\widetilde{\mathbf{\uptau}}}_r)^{T}\Gamma^{-1}\mathbf{Q}$. When $\mathbf{K}_1$, $\mathbf{K}_2$ and $\Gamma$ are assigned with large enough values at the design step, $\mathbf{\dot{V}}\leq 0$ can be achieved and $\mathbf{V}$ is ensured to be bounded, thus leading to the conclusion that $\mathbf{Q}$ is bounded. Then design a new Lyapunov function as
\begin{equation}
    \mathbf{V}_2=\frac{1}{4\lambda}\mathbf{s}^{T}\mathbf{M}\mathbf{s},
\end{equation}
whose derivative can be deducted as,
\begin{equation}
    \mathbf{\dot{V}}_2=\mathbf{s}^{T}(\mathbf{Q}-\mathbf{K}_1\mathbf{s}-\mathbf{K}_2|s|^{r}sign(s)),
\end{equation}
where  $0< r <1$. Suppose $||Q||< a$,
\begin{equation}
    \mathbf{\dot{V}}_2\leq \frac{1}{2}||s||^2+\frac{1}{2}a-\lambda_{min}(\mathbf{K}_1)||s||^2-\lambda_{min}(\mathbf{K}_2)||s||^{1+r},
\end{equation}
choose $\mathbf{K}_1$ when $\lambda_{min}(\mathbf{K}_1)> \frac{1}{2}+\beta$, where $\beta > 0$,
\begin{equation}
    \mathbf{\dot{V}}_2\leq -\beta||s||^2-\lambda_{min}(\mathbf{K}_2)||s||^{1+r}+\frac{1}{2}a,
\end{equation}
which induces that the Lyapunov function converges to a range close to zero in a finite time and $\mathbf{s}$ converges to a range close to zero in a finite time. Therefore, the supposed condition of the Lyapunov theorem can be regarded as satisfied, thus proving the stability of the designed SMC. 



\ifCLASSOPTIONcaptionsoff
  \newpage
\fi



%

\scriptsize
\bibliographystyle{IEEEtran}
\bibliography{ref}

\begin{thebibliography}{10}
\providecommand{\url}[1]{#1}
\csname url@samestyle\endcsname
\providecommand{\newblock}{\relax}
\providecommand{\bibinfo}[2]{#2}
\providecommand{\BIBentrySTDinterwordspacing}{\spaceskip=0pt\relax}
\providecommand{\BIBentryALTinterwordstretchfactor}{4}
\providecommand{\BIBentryALTinterwordspacing}{\spaceskip=\fontdimen2\font plus
\BIBentryALTinterwordstretchfactor\fontdimen3\font minus
  \fontdimen4\font\relax}
\providecommand{\BIBforeignlanguage}[2]{{%
\expandafter\ifx\csname l@#1\endcsname\relax
\typeout{** WARNING: IEEEtran.bst: No hyphenation pattern has been}%
\typeout{** loaded for the language `#1'. Using the pattern for}%
\typeout{** the default language instead.}%
\else
\language=\csname l@#1\endcsname
\fi
#2}}
\providecommand{\BIBdecl}{\relax}
\BIBdecl

\bibitem{Baran2018}
T.~Baraniuk, R.~Simoni, and L.~Weihmann,
  ``\BIBforeignlanguage{English}{Fault-tolerant architecture for {AUV}s},'' in
  \emph{\BIBforeignlanguage{English}{Proceedings of 2018 IEEE/OES Autonomous
  Underwater Vehicle Workshop}}, Porto, Portugal, 2018, pp. 1--6.

\bibitem{Chaosteerwire2019}
C.~Huang, F.~Naghdy, H.~Du, and H.~Huang, ``\BIBforeignlanguage{English}{Fault
  tolerant steer-by-wire systems: An overview},''
  \emph{\BIBforeignlanguage{English}{Annual Reviews in Control}}, vol.~47, pp.
  98--111, 2019.

\bibitem{xin2014}
X.~Qi, J.~Qi, D.~Theilliol, Y.~Zhang, J.~Han, D.~Song, and C.~Hua,
  ``\BIBforeignlanguage{English}{A review on fault diagnosis and fault tolerant
  control methods for single-rotor aerial vehicles},''
  \emph{\BIBforeignlanguage{English}{J. Intell. Robot. Syst.}}, vol.~73, no.
  1-4, pp. 535--555, Jan. 2014.

\bibitem{Lu2016}
K.~Lu, Y.~Xia, C.~Yu, and H.~Liu, ``\BIBforeignlanguage{English}{Finite-time
  tracking control of rigid spacecraft under actuator saturations and
  faults},'' \emph{\BIBforeignlanguage{English}{IEEE Trans. Autom. Sci. Eng.}},
  vol.~13, no.~1, pp. 368--381, Jan. 2016.

\bibitem{TITS_mao2020}
Z.~Mao, X.~Yan, B.~Jiang, and M.~Chen, ``\BIBforeignlanguage{English}{Adaptive
  fault-tolerant sliding-mode control for high-speed trains with actuator
  faults and uncertainties},'' \emph{\BIBforeignlanguage{English}{IEEE Trans.
  Intell. Transp. Syst.}}, vol.~21, no.~6, pp. 2449--2460, Jul. 2020.

\bibitem{Meyer2018}
R.~T. Meyer, S.~C. Johnson, R.~A. DeCarlo, S.~Pekarek, and S.~D. Sudhoff,
  ``\BIBforeignlanguage{English}{Hybrid electric vehicle fault tolerant
  control},'' \emph{\BIBforeignlanguage{English}{J. Dyn. Syst-T. ASME}}, vol.
  140, no.~2, pp. 1--12, 2018.

\bibitem{xiong2018}
R.~Xiong, Q.~Yu, and W.~Shen, ``\BIBforeignlanguage{English}{Review on sensors
  fault diagnosis and fault-tolerant techniques for lithium ion batteries in
  electric vehicles},'' in \emph{\BIBforeignlanguage{English}{Proceedings of
  2018 IEEE Conference on Industrial Electronics and Applications (ICIEA)}},
  Wuhan, China, 2018, pp. 406--410.

\bibitem{gong2018tec1}
J.~Gong and M.~Yang, ``\BIBforeignlanguage{English}{Evolutionary fault
  tolerance method based on virtual reconfigurable circuit with neural network
  architecture},'' \emph{\BIBforeignlanguage{English}{IEEE Transactions on
  Evolutionary Computation}}, vol.~22, no.~6, pp. 949--960, Dec. 2018.

\bibitem{motorsCST1}
G.~Zhang, H.~Zhang, X.~Huang, J.~Wang, H.~Yu, and R.~Graaf,
  ``\BIBforeignlanguage{English}{Active fault-tolerant control for electric
  vehicles with independently driven rear in-wheel motors against certain
  actuator faults},'' \emph{\BIBforeignlanguage{English}{IEEE Trans. Control
  Syst. Technol.}}, vol.~24, no.~5, pp. 1557--1572, 2016.

\bibitem{Lang2021}
X.~Lang, T.~Yang, Z.~Wang, C.~Wang, S.~Bozhko, and P.~Wheeler,
  ``\BIBforeignlanguage{English}{Fault tolerant control of advanced power
  generation center for more-electric aircraft applications},''
  \emph{\BIBforeignlanguage{English}{IEEE Trans. Transp. Electrification}}, p.
  Article in press. DOI: dx.doi.org/10.1109/TTE.2021.3093506, 2021.

\bibitem{Cao2021}
X.~Cao, Y.~Tian, X.~Ji, and B.~Qiu,
  ``\BIBforeignlanguage{English}{Fault-tolerant controller design for path
  following of the autonomous vehicle under the faults in braking actuators},''
  \emph{\BIBforeignlanguage{English}{IEEE Trans. Transp. Electrification}},
  vol.~7, no.~4, pp. 2530--2540, 2021.

\bibitem{Seto2020}
M.~Seto and K.~Svendsen, \emph{\BIBforeignlanguage{English}{Autonomous
  Underwater Vehicles: Design and practice}}.\hskip 1em plus 0.5em minus
  0.4em\relax The Institution of Engineering and Technology, 2020, ch. Advanced
  {AUV} fault management, pp. 419--445.

\bibitem{ocftc1}
J.~Kadiyam, A.~Parashar, S.~Mohan, and D.~Deshmukh,
  ``\BIBforeignlanguage{English}{Actuator fault-tolerant control study of an
  underwater robot with four rotatable thrusters},''
  \emph{\BIBforeignlanguage{English}{Ocean Eng.}}, vol. 197, pp. 961--979,
  2020.

\bibitem{TITS1_MSV2021}
G.~Zhu, Y.~Ma, Z.~Li, R.~Malekian, and M.~Sotelo,
  ``\BIBforeignlanguage{English}{Event-triggered adaptive neural fault-tolerant
  control of underactuated msvs with input saturation},''
  \emph{\BIBforeignlanguage{English}{IEEE Trans. Intell. Transp. Syst.}},
  vol.~23, no.~7, pp. 7045--7057, 2022.

\bibitem{UUV1}
J.~Li, Z.~Xu, D.~Zhu, K.~Dong, Z.~Z. Tao~Yan, and S.~X. Yang,
  ``\BIBforeignlanguage{English}{Bio-inspired intelligence with applications to
  robotics: a survey},'' \emph{\BIBforeignlanguage{English}{Intelligence \&
  Robotics}}, vol.~1, no.~1, pp. 58--83, 2021.

\bibitem{TSMC1}
T.~Li, G.~Li, and Q.~Zhao, ``\BIBforeignlanguage{English}{Adaptive
  fault-tolerant stochastic shape control with application to particle
  distribution control},'' \emph{\BIBforeignlanguage{English}{IEEE Trans. Syst.
  Man Cybern., Syst.}}, vol.~45, no.~12, pp. 1592--1604, Dec. 2015.

\bibitem{martynova2020}
L.~Martynova and M.~Rozengauz, ``\BIBforeignlanguage{English}{Approach to
  reconfiguration of a motion control system for an autonomous underwater
  vehicle},'' \emph{\BIBforeignlanguage{English}{Gyroscopy and Navigation}},
  vol.~11, no.~3, pp. 244--253, 2020.

\bibitem{ocftc2}
X.~Liu, M.~Zhang, and F.~Yao, ``\BIBforeignlanguage{English}{Adaptive fault
  tolerant control and thruster fault reconstruction for autonomous underwater
  vehicle},'' \emph{\BIBforeignlanguage{English}{Ocean Eng.}}, vol. 155, pp. 10
  -- 23, May 2018.

\bibitem{Omerdic2004}
E.~Omerdic and G.~Roberts, ``\BIBforeignlanguage{English}{Thruster fault
  diagnosis and accommodation for open-frame underwater vehicles},''
  \emph{\BIBforeignlanguage{English}{Control Eng. Pract.}}, vol.~12, no.~12,
  pp. 1575--1598, Dec. 2004.

\bibitem{Podder2000}
T.~Podder, G.~Antonelli, and N.~Sarkar, ``\BIBforeignlanguage{English}{Fault
  tolerant control of an autonomous underwater vehicle under thruster
  redundancy: simulations and experiments},'' in
  \emph{\BIBforeignlanguage{English}{Proceedings of 2000 IEEE International
  Conference on Robotics and Automation (ICRA)}}, vol.~2, San Francisco, CA,
  USA, 2000, pp. 1251--1256.

\bibitem{Zhang2021}
S.~Zhang, Y.~Wu, X.~He, and Z.~Liu, ``\BIBforeignlanguage{English}{Cooperative
  fault-tolerant control for a mobile dual flexible manipulator with output
  constraints},'' \emph{\BIBforeignlanguage{English}{IEEE Trans. Autom. Sci.
  Eng.}}, 2021.

\bibitem{Shen2018}
C.~Shen, Y.~Shi, and B.~Buckham, ``\BIBforeignlanguage{English}{Trajectory
  tracking control of an autonomous underwater vehicle using lyapunov-based
  model predictive control},'' \emph{\BIBforeignlanguage{English}{IEEE Trans.
  Ind. Electron.}}, vol.~65, no.~7, pp. 5796--5805, 2018.

\bibitem{TSMC2}
Q.~Wen, R.~Kumar, and J.~Huang, ``\BIBforeignlanguage{English}{Framework for
  optimal fault-tolerant control synthesis: maximize prefault while minimize
  post-fault behaviors},'' \emph{\BIBforeignlanguage{English}{IEEE Trans. Syst.
  Man Cybern., Syst.}}, vol.~44, no.~8, pp. 1056--1066, Aug. 2014.

\bibitem{UUV2}
D.~Zhu, T.~Yan, and S.~X. Yang, ``\BIBforeignlanguage{English}{Motion planning
  and tracking control of unmanned underwater vehicles: technologies,
  challenges and prospects},'' \emph{\BIBforeignlanguage{English}{Intelligence
  \& Robotics}}, vol.~2, no.~3, pp. 200--222, 2022.

\bibitem{Sinha2020}
U.~Sinha, M.~Cashman, and M.~Cohen, ``\BIBforeignlanguage{English}{Using a
  genetic algorithm to optimize configurations in a data-driven application},''
  in \emph{\BIBforeignlanguage{English}{Proceedings of 12th International
  Symposium on Search-Based Software Engineering}}, Cham, Switzerland, 2020,
  pp. 137--152.

\bibitem{mohanty2020}
S.~Mohanty and S.~Mohanty, ``\BIBforeignlanguage{English}{Genetic
  algorithm-based motif search problem: a review},'' in
  \emph{\BIBforeignlanguage{English}{Proceedings of 2020 3rd International
  Conference on Smart Computing and Informatics. Smart Innovation, Systems and
  Technologies (SIST)}}, vol.~1, Singapore, Singapore, 2020, pp. 719--731.

\bibitem{dunweitec2}
D.~Gong, J.~Sun, and Z.~Miao, ``\BIBforeignlanguage{English}{A set-based
  genetic algorithm for interval many-objective optimization problems},''
  \emph{\BIBforeignlanguage{English}{IEEE Trans. Evol. Comput.}}, vol.~22,
  no.~1, pp. 47--60, Feb. 2018.

\bibitem{dataconvNNChi2019}
C.~Lai, H.~Ibrahim, M.~Abdullah, J.~Abdullah, S.~Suandi, and A.~Azman,
  ``\BIBforeignlanguage{English}{A literature review on data conversion methods
  on {EEG} for convolution neural network applications},'' in
  \emph{\BIBforeignlanguage{English}{Proceedings of 2019 International
  Conference on Robotics, Vision, Signal Processing and Power Applications.
  Enabling Research and Innovation Towards Sustainability. Lecture Notes in
  Electrical Engineering (LNEE)}}, Singapore, Singapore, 2019, pp. 521--527.

\bibitem{NNAsmara2020}
A.~Asmara, ``\BIBforeignlanguage{English}{A possibility of artificial neural
  networks to be applied in the predictive test: a systematic literature review
  and study case},'' \emph{\BIBforeignlanguage{English}{J. Phys., Conf. Ser.}},
  vol. 1456, pp. 12\,052--12\,061, 2020.

\bibitem{greedyHul2017}
S.~Hulyalkar and H.~Khanuja, ``\BIBforeignlanguage{English}{Optimal solution
  generation from reviews and micro-reviews using greedy algorithm},'' in
  \emph{\BIBforeignlanguage{English}{Proceedings of 2017 International
  Conference on Computing, Communication, Control and Automation (ICCUBEA)}},
  Pune, India, 2017, pp. 1--6.

\bibitem{Guo2018}
P.~Guo, M.~Liu, and Z.~Xue, ``\BIBforeignlanguage{English}{A {PSO}-based
  energy-efficient fault-tolerant static scheduling algorithm for real-time
  tasks in clouds},'' in \emph{\BIBforeignlanguage{English}{Proceedings of 4th
  International Conference on Computer and Communications (ICCC)}}, Chengdu,
  China, 2018, pp. 2537--2541.

\bibitem{Celtek2020}
S.~Celtek, A.~Durdu, and M.~Ali, ``Real-time traffic signal control with swarm
  optimization methods,'' \emph{Measurement}, vol. 166, pp. 108\,206--108\,213,
  2020.

\bibitem{TITSwu2021_pso}
J.~Wu, C.~Song, J.~Ma, J.~Wu, and G.~Han,
  ``\BIBforeignlanguage{English}{Reinforcement learning and particle swarm
  optimization supporting real-time rescue assignments for multiple autonomous
  underwater vehicles},'' \emph{\BIBforeignlanguage{English}{IEEE Trans.
  Intell. Transp. Syst.}}, vol.~23, pp. 6807--6820, Jul. 2021.

\bibitem{Liu2011}
D.~Zhu, Q.~Liu, and Z.~Hu, ``\BIBforeignlanguage{English}{Fault-tolerant
  control algorithm of the manned submarine with multi-thruster based on
  quantum-behaved particle swarm optimisation},''
  \emph{\BIBforeignlanguage{English}{Int. J. Control}}, vol.~84, no.~11, pp.
  1817--1829, Nov. 2011.

\bibitem{Guo2015}
W.~Guo, J.~Li, G.~Chen, Y.~Niu, and C.~Chen, ``\BIBforeignlanguage{English}{A
  {PSO}-optimized real-time fault-tolerant task allocation algorithm in
  wireless sensor networks},'' \emph{\BIBforeignlanguage{English}{IEEE Trans.
  Parall. Distr. Syst.}}, vol.~26, no.~12, pp. 3236--3249, 2015.

\bibitem{Saremi2017}
S.~Saremi, S.~Mirjalili, and A.~Lewis, ``Grasshopper optimisation algorithm:
  theory and application,'' \emph{Adv. Eng. Softw.}, vol. 105, pp. 30--47,
  2017.

\bibitem{Guo2021}
G.~Wang, A.~Heidari, M.~Wang, F.~Kuang, W.~Zhu, and H.~Chen,
  ``\BIBforeignlanguage{English}{Chaotic arc adaptive grasshopper
  optimization},'' \emph{\BIBforeignlanguage{English}{IEEE Access}}, vol.~9,
  pp. 17\,672--17\,706, 2021.

\bibitem{Xia2021}
J.~Xia, Z.~Wang, D.~Yang, R.~Li, G.~Liang, H.~Chen, A.~A. Heidari, H.~Turabieh,
  M.~Mafarja, and Z.~Pan, ``\BIBforeignlanguage{English}{Performance
  optimization of support vector machine with oppositional grasshopper
  optimization for acute appendicitis diagnosis},''
  \emph{\BIBforeignlanguage{English}{Computers in Biology and Medicine}}, vol.
  143, p. Article in press. DOI: dx.doi.org/10.1016/j.compbiomed.2021.105206,
  2022.

\bibitem{jianfa2017}
J.~Wu, H.~Wang, N.~Li, P.~Yao, Y.~Huang, Z.~Su, and Y.~Yu,
  ``\BIBforeignlanguage{English}{Distributed trajectory optimization for
  multiple solar-powered {UAV}s target tracking in urban environment by
  adaptive grasshopper optimization algorithm},''
  \emph{\BIBforeignlanguage{English}{Aerosp. Sci. Technol.}}, vol.~70, pp.
  497--510, Nov. 2017.

\bibitem{Mishra2020}
P.~Mishra, V.~Goyal, and A.~Shukla, \emph{\BIBforeignlanguage{English}{Advances
  in Intelligent Computing and Communication. Lecture Notes in Networks and
  Systems}}.\hskip 1em plus 0.5em minus 0.4em\relax Springer, Singapore, 2020,
  ch. An Improved Grasshopper Optimization Algorithm for Solving Numerical
  Optimization Problems, pp. 179--188.

\bibitem{Durham1993}
W.~Durham, ``\BIBforeignlanguage{English}{Constrained control allocation},''
  \emph{\BIBforeignlanguage{English}{J. Guid. Control Dyn.}}, vol.~16, no.~4,
  pp. 717--725, Jul. 1993.

\bibitem{r28}
D.~Zhu and B.~Sun, ``\BIBforeignlanguage{English}{The bio-inspired model based
  hybrid sliding-mode tracking control for unmanned underwater vehicles},''
  \emph{\BIBforeignlanguage{English}{Eng. Appl. Artif. Intell.}}, vol.~26,
  no.~10, pp. 2260--2269, Nov. 2013.

\bibitem{r29}
Y.~Shtessel, D.~Foreman, and C.~Tournes,
  ``\BIBforeignlanguage{English}{Stability margins in traditional and second
  order sliding mode control},'' in
  \emph{\BIBforeignlanguage{English}{Proceedings of 2011 IEEE Conference on
  Decision and Control}}, Orlando, FL, USA, 2011, pp. 4604--4609.

\bibitem{r25}
J.~Zand, ``\BIBforeignlanguage{English}{Enhanced navigation and tether
  management of inspection class remotely operated vehicles},'' Master Thesis,
  Mechanical Engineering, University of Victoria, Victoria, Canada, 2005.

\bibitem{danjie2021OE}
D.~Zhu, L.~Wang, Z.~Hu, and S.~X. Yang, ``\BIBforeignlanguage{English}{A
  grasshopper optimization-based fault-tolerant control algorithm for a human
  occupied submarine with the multi-thruster system},''
  \emph{\BIBforeignlanguage{English}{Ocean Eng.}}, vol. 242, 2021.

\bibitem{swarming1}
S.~J. Simpson, A.~R. Mccaffery, and B.~F. HÄGELE, ``A behavioural analysis of
  phase change in the desert locust,'' \emph{Biological Reviews}, vol.~74,
  no.~4, pp. 461--480, 2010.

\bibitem{swarming2}
S.~M. Rogers, T.~Matheson, E.~Despland, T.~Dodgson, and S.~J. Simpson,
  ``Mechanosensory-induced behavioral gregarization in the desert locust
  schistocerca gregaria,'' \emph{Journal of Experimental Biology}, vol. 206,
  no. Pt 22, pp. 3991--4002, 2003.

\bibitem{Meraihi2021}
Y.~Meraihi, A.~Gabis, S.~Mirjalili, and A.~Ramdane-Cherif,
  ``\BIBforeignlanguage{English}{Grasshopper optimization algorithm: theory,
  variants, and applications},'' \emph{\BIBforeignlanguage{English}{IEEE
  Access}}, vol.~9, pp. 50\,001--50\,024, 2021.

\bibitem{wave1}
Y.~Zhao, X.~Qi, Y.~Ma, Z.~Li, R.~Malekian, and M.~Sotelo,
  ``\BIBforeignlanguage{English}{Path following optimization for an
  underactuated {USV} using smoothly-convergent deep reinforcement learning},''
  \emph{\BIBforeignlanguage{English}{IEEE Trans. Intell. Transp. Syst.}},
  vol.~22, no.~10, pp. 6208--6220, 2021.

\bibitem{wave2}
Z.~Peng, D.~Wang, Z.~Chen, X.~Hu, and W.~Lan,
  ``\BIBforeignlanguage{English}{Adaptive dynamic surface control for
  formations of autonomous surface vehicles with uncertain dynamics},''
  \emph{\BIBforeignlanguage{English}{IEEE Trans. Control Syst. Technol.}},
  vol.~21, no.~2, pp. 513--520, 2013.

\bibitem{wave3}
P.~Gong, Z.~Yan, W.~Zhang, and J.~Tang,
  ``\BIBforeignlanguage{English}{Trajectory tracking control for autonomous
  underwater vehicles based on dual closed-loop of mpc with uncertain
  dynamics},'' \emph{\BIBforeignlanguage{English}{Ocean Eng.}}, vol. 265, 2022.

\end{thebibliography}

\end{document}